\def\PublicPreprint{1}
\def\PaperSystem{RankLane}
\def\PaperTitle{RankLane: Efficient MoE Prefill on\texorpdfstring{\\}{ }Accelerators with Ranked Expert Capacities}
\def\PaperRunningTitle{RankLane: Efficient MoE Prefill on Accelerators}
\newcommand{\PaperAuthorBlock}{%
  \begin{mlsysauthorlist}
  \mlsysauthor{Wentao Ni}{ucsd}
  \mlsysauthor{Chihao Yu}{ucsd}
  \mlsysauthor{Hong Cai}{qualcomm}
  \mlsysauthor{Mingu Lee}{qualcomm}
  \mlsysauthor{Hejia Zhang}{ucsd}
  \mlsysauthor{Fatih Porikli}{qualcomm}
  \mlsysauthor{Jishen Zhao}{ucsd}
  \end{mlsysauthorlist}
  \mlsysaffiliation{ucsd}{University of California, San Diego}
  \mlsysaffiliation{qualcomm}{Qualcomm AI Research}
  \mlsyscorrespondingauthor{Jishen Zhao}{jzhao@ucsd.edu}
}

\documentclass{article}

\usepackage[T1]{fontenc}
\usepackage{microtype}
\usepackage{graphicx}
\usepackage{subfigure}
\usepackage{booktabs}
\usepackage{array}
\usepackage{float}
\usepackage{amsmath}
\usepackage{amssymb}
\usepackage{xspace}
\usepackage{tikz}
\usetikzlibrary{arrows.meta,calc}
\usepackage{hyperref}

\ifdefined\PublicPreprint
  \usepackage[accepted]{mlsys2025}
  \makeatletter
  \renewcommand{\Notice@String}{Preprint.}
  \makeatother
  \hypersetup{pdfsubject={\PaperSystem\space preprint}}
\else
  \usepackage{mlsys2025}
  \hypersetup{pdfsubject={MLSys 2027 submission}}
\fi

\newcommand{\sys}{\PaperSystem\xspace}

\mlsystitlerunning{\PaperRunningTitle}

\begin{document}

\twocolumn[
\mlsystitle{\PaperTitle}

\ifdefined\PublicPreprint
  \PaperAuthorBlock
\else
  \begin{mlsysauthorlist}
  \mlsysauthor{Anonymous Authors}{anon}
  \end{mlsysauthorlist}
  \mlsysaffiliation{anon}{Anonymous Institution}
  \mlsyscorrespondingauthor{Anonymous Authors}{anonymous@example.com}
\fi

\mlsyskeywords{Mixture of Experts, LLM Inference, Static Execution Plans, Inference Accelerators}

\vskip 0.3in

\begin{abstract}
Mixture-of-experts (MoE) models expand model capacity by activating only a subset of experts for each token.
On accelerators with static execution plans, expert capacities and placement are fixed before routing reveals each
expert's load. Uniform worst-case capacities waste work on padding; smaller capacities tied to expert identities can overflow.
We present \sys, which ranks each device's experts by current load and assigns them to fixed-capacity lanes. Runtime
indices select experts and token rows while tier populations, shapes and device placement remain fixed. An aggregation
graph preserves device ownership and reduces each token's local contributions on one core. Routing calibration
and measured execution costs guide tier selection; overflowing chunks rerun at full capacity. We evaluate full-model
prefill of Qwen3-30B-A3B and Gemma 4 26B-A4B with MXFP6 weights on four Cloud AI 100 Ultra SoCs, using
128--2048-token chunks and including overflow detection and in-program recovery support. Against a dense baseline,
\sys achieves end-to-end prefill speedups of
1.27--4.18$\times$ on Qwen3 and 1.93--3.46$\times$ on Gemma.
\end{abstract}

]

\printAffiliationsAndNotice{}

\raggedbottom

\section{Introduction}
\label{sec:intro}

Mixture-of-experts (MoE) models expand model capacity while activating only a subset of experts for each token
\citep{shazeer2017outrageously,fedus2022switch}. Efficient prefill matters for serving these models because the prompt
must be processed before generation begins. Chunked prefill bounds the tokens processed per invocation
\citep{agrawal2024sarathi}, but each expert's load still varies with the input and layer. Specialized inference
accelerators target high performance per watt and lower serving cost \citep{chatha2021qualcomm,awsinf2}.
Turning MoE's sparse computation into efficient execution on these platforms requires accommodating this changing demand.

\begin{figure*}[t]
\centering
\includegraphics[width=\textwidth]{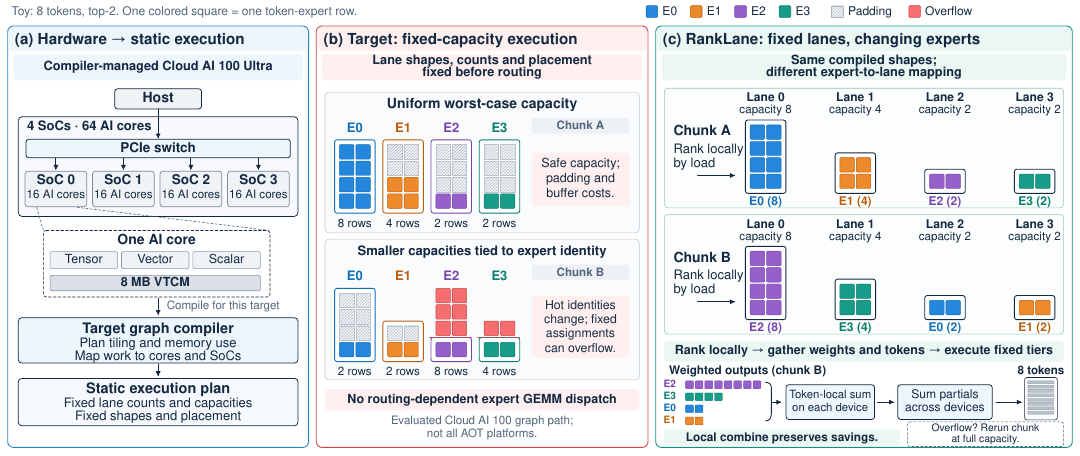}
\caption{Static execution and changing expert demand. (a) The evaluated Cloud AI 100 graph path fixes lane capacities,
shapes and placement. (b) For eight top-2 tokens on one device, uniform capacity reserves 32 rows for 16 assignments;
smaller capacities tied to expert identities overflow when the busy experts change. (c) Ranking fits both illustrated
chunks into the same 16-row lane layout, followed by local aggregation.}
\label{fig:static-motivation}
\end{figure*}

This becomes difficult when the accelerator executes a static plan that fixes tensor shapes, expert capacities and
placement before routing. We study this setting on Qualcomm Cloud AI 100, where fixed-capacity \emph{lanes} execute
expert GEMMs (Figure~\ref{fig:static-motivation}(a)). GPU block-sparse kernels adapt their work to routed tokens
\citep{gale2023megablocks}; the evaluated graph interface instead executes expert GEMMs in slots with compiled
capacities. Runtime indices select weights and tokens within these slots; GEMM shapes and dispatch remain fixed.

Two coupled challenges limit efficiency. First, capacities tied to expert identities must accommodate their changing
peak loads. Qwen3-30B-A3B selects 8 of 128 experts per token \citep{yang2025qwen3}: capacity $T$ per expert covers
any routing of a $T$-token chunk but reserves $128T$ rows for $8T$ assignments. Smaller fixed capacities risk overflow
when the busy experts change (Figure~\ref{fig:static-motivation}(b)). Second, returning expert outputs to tokens can
erase the savings. Dense aggregation retains large lane-indexed buffers, and compiler placement can redistribute
outputs between devices and cores. The latency benefit of smaller expert capacities therefore depends on the storage
and locality of the return path (\S\ref{sec:mot}).

We present \sys, which assigns capacities by current load rank while preserving a static execution plan
(Figure~\ref{fig:static-motivation}(c)). Our key observation is that the busiest experts change across inputs,
while the measured load envelope falls sharply below the first ranks. Ranking each device's experts and assigning
fixed rank intervals to tiers gives the largest lanes to the currently busiest experts. Runtime gathers select their
weights and token rows while tier populations, tensor shapes and expert-to-device placement remain fixed.
To realize these capacity savings, token-owned aggregation gathers a token's local expert outputs and sums them
on one core within each device, limiting intermediate storage and communication.

The paper makes three contributions. First, we characterize the capacity savings available from runtime load ranks
and show how aggregation governs their latency benefit (\S\ref{sec:mot}). Second, we design a compiled program
that combines adaptive expert assignment with device-local, token-owned reduction (\S\ref{sec:design}). Third,
we integrate calibration and measured execution costs to select tiers, with overflow detection and capacity-$T$
reruns for loads beyond calibration. Together, these mechanisms preserve token--expert assignments and keep the
normal path within the compiled program, without host dispatch between layers.

We evaluate full-model prefill of Qwen3-30B-A3B and Gemma 4 26B-A4B with MXFP6 weights and an on-device KV cache
on four Cloud AI 100 Ultra SoCs. Matched ablations isolate the contributions of tiering and reduction, and held-out
routing and recovery experiments test capacity coverage (\S\ref{sec:eval}). Across 128--2048-token chunks,
\sys includes overflow detection and in-program recovery support and achieves end-to-end prefill speedups over a
dense baseline of 1.27--4.18$\times$ on Qwen3 and 1.93--3.46$\times$ on Gemma.

\section{Background and Related Work}
\label{sec:bg}

\subsection{MoE Inference}

An MoE layer replaces the feed-forward block with $E$ experts and a router. For a token with hidden state
$x_t\in\mathbb{R}^d$, the router selects the set $\mathcal{K}_t$ of $k$ experts with the largest gate values and returns
\begin{equation}
y_t=\sum_{e\in\mathcal{K}_t} w_{t,e}\,W^{\mathrm{down}}_e\big(\mathrm{SiLU}(W^{\mathrm{gate}}_e x_t)\odot W^{\mathrm{up}}_e x_t\big),
\label{eq:moe}
\end{equation}
with normalized gate weights $w_{t,e}$. Qwen3-30B-A3B uses $E=128$, $k=8$, $d=2048$ and an expert width of 768, so an
expert's three matrices hold 9\,MiB in FP16. Implementations group assignments by expert, run each expert's GEMMs over
its gathered token rows, and return the weighted rows to their tokens; under expert parallelism the experts are spread
over devices and dispatch and combine become cross-device exchanges \citep{lepikhin2021gshard,rajbhandari2022deepspeedmoe}.
Training systems bound each expert's work by a capacity factor and drop the tokens beyond it
\citep{lepikhin2021gshard,fedus2022switch}; at inference, a dropped assignment changes the model's output. Prefill
processes the prompt in chunks of $T$ tokens and writes their keys and values to the KV cache; decode then generates one
token per step \citep{kwon2023vllm,agrawal2024sarathi}. A chunk gives each layer $8T$ assignments, $2T$ per SoC on average
when four SoCs hold 32 experts each.

\subsection{MoE Systems}

MegaBlocks represents routed expert work as a runtime block-sparse topology and executes it with GPU kernels
without dropping tokens \citep{gale2023megablocks}. DeepSpeed-MoE, FasterMoE and Lina optimize or schedule the
all-to-all exchange \citep{rajbhandari2022deepspeedmoe,he2022fastermoe,li2023lina}. The combine has been made sparse on
GPUs too: Tutel replaces GShard's dense dispatch and combine einsums with sparse encode and decode kernels that assign
tokens to thread groups \citep{hwang2023tutel}, and vLLM sums each token's top-$k$
expert rows in a single kernel, with fixed-size buffers and device-side indices that suit CUDA graphs
\citep{vllmmoe,megatronfusedmoe}. A kernel author decides there which thread reads which rows and where partial sums live;
a graph compiler decides that for the program, which is the gap \S\ref{sec:mot-red} examines. Placement studies exploit routing structure
\citep{yao2024exflow,yu2025patterns}, \citet{huang2024toward} replace static capacity with dynamic gating, and
expert-choice routing fixes each expert's tokens by construction but changes the model \citep{zhou2022expertchoice}. On
Trainium, expert work is packed into fixed-size blocks whose weight loads are shared across an expert's blocks and whose
unused blocks are skipped \citep{awsneuronmoe}. DynaMoE recompiles a
static-shape program when capacities change during training \citep{kossmann2022dynamoe}; our full-model programs take
7--26 minutes to compile, so capacities cannot follow each chunk by recompilation.

NPUMoE targets Apple Silicon NPUs, assigning experts to static capacity tiers using offline popularity and performing
routing and aggregation on the CPU \citep{benazir2026npumoe}. \sys assigns experts to tiers by current load rank
and expresses assignment and aggregation within the compiled multi-device program.

\subsection{Accelerators with Static Execution Plans}

Groq statically schedules instructions and data movement \citep{abts2020thinkfast}; Neuron compiles fixed-shape graphs
for traced inference on Inferentia and Trainium \citep{awsneurontrace}. Compiler-managed graphs also appear in
SambaNova's RDU and Graphcore's IPU. SambaFlow maps dataflow graphs onto RDUs
and produces a compiled executable \citep{sambanova2024software}; Poplar compiles static graphs with tensor dimensions
determined before execution \citep{graphcoreinference}. Compilation ahead of time does not itself require fixed
expert capacities: runtime control flow and custom-kernel interfaces vary across platforms. \sys targets the constraints
in \S\ref{sec:intro}; transfer requires cost and locality checks (Appendix~\ref{app:transfer}).

A Cloud AI 100 SoC has 16 AI cores and 136\,GB/s of LPDDR4x bandwidth to up to 32\,GB
\citep{chatha2021qualcomm,qualcomm2025arch}; each core has a scalar VLIW processor, a vector unit, a tensor unit with 4K
FP16 MACs and 8\,MB of software-managed vector memory. The Ultra card carries four SoCs under a shared 150\,W limit.

The compiler takes an ONNX graph and one shape specialization per chunk length and produces a static program,
partitioned over the cores of a SoC and, by multi-device tensor slicing, over the SoCs. Matrix weights can be MXFP6
\citep{rouhani2023mx}, dequantized by the vector unit while the tensor unit computes. A KV cache stays on device when each
cache input can be paired with its updated output (\emph{retained state}). The evaluated graph interface supports
runtime weight gathers but rejects ONNX \texttt{If} with a non-constant condition. Dynamic-shape compilers for CPUs and GPUs
provide run-time dispatch \citep{shen2021nimble,zhu2021disc}; here, expert GEMMs execute in slots with compiled capacities.

\section{Motivation}
\label{sec:mot}

Under this execution model, expert capacity is allocated before routing reveals demand. We study this
problem on Qwen3-30B-A3B compiled for the four SoCs of one Cloud AI 100 Ultra card (setup in \S\ref{sec:setup}). The
exported graph \citep{qefficient} runs a layer's 128 experts as two stages of 64 \emph{lanes}: each lane executes one
expert's gate, up and down GEMMs over a fixed number of token rows, its capacity $C$. Compilation fixes the lanes'
shapes, placement and schedule. Figure~\ref{fig:static-motivation} connects this hardware setting to the capacity
mismatch and the opportunity to adapt expert assignments within the compiled plan. We first examine the capacity
this can save, then the aggregation needed to realize the saving.

\subsection{Fixed Capacities for Runtime Load Ranks}
\label{sec:mot-pad}

A fixed assignment binds a lane's capacity to one expert. Capacity $T$ accommodates any routing of a $T$-token chunk,
but reserves $128T$ rows for Qwen3's $8T$ assignments. Smaller capacities must cover each expert's peak demand over the
inputs the program will serve; average popularity alone does not determine that peak. Figure~\ref{fig:static-motivation}(b--c)
shows four experts with token counts $(8,4,2,2)$ in one chunk and $(2,2,8,4)$ in another. Fixed expert assignments require
capacities $(8,4,8,4)$ to cover both. Assigning lanes by current load rank requires $(8,4,2,2)$ in either chunk: the
expert needing the largest lane changes, while the lane capacities stay fixed.

We measure this opportunity using routing captures of all 48 layers for 50 prompts of 128 tokens from each of eight
workloads, GSM8K, MMLU,
SWE-bench Lite issues, HumanEval and its JavaScript version, Alpaca instructions, CNN/DailyMail articles and Chinese XNLI
sentence pairs \citep{cobbe2021gsm8k,hendrycks2021mmlu,jimenez2024swebench,chen2021humaneval,muennighoff2024octopack,taori2023alpaca,hermann2015cnndm,conneau2018xnli},
and for chunks of 256 and 512 tokens from four of them (Appendix~\ref{app:calib}). An expert in some layer receives
127 of 128, 256 of 256 or 508 of 512 tokens. Which experts are busy also varies: across workloads the 16 most loaded
experts of a layer agree with a Jaccard index of 0.00--0.33, against 0.52--0.88 between two halves of one workload.

\begin{figure}[t]
\centering
\includegraphics[width=\columnwidth]{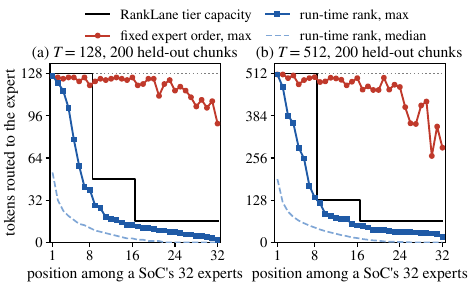}
\vskip -0.08in
\caption{Expert-load envelopes from device routing captures. Each panel uses 200 calibration and 200 held-out chunks,
with matched counts per workload across eight workloads. Calibration fixes the mean-load expert order; curves summarize held-out loads.
Red: fixed order. Blue: runtime rank. Black: deployed \sys capacities. Dotted: capacity $T$.}
\label{fig:rank}
\vskip -0.1in
\end{figure}

Figure~\ref{fig:rank} compares the capacity envelopes using device-captured routing. With each SoC's 32 experts fixed in order of their calibration
mean load, the held-out maximum is 90--126 of 128 rows at every position; covering this envelope reserves 93\% of the
rows of capacity-$T$ padding. With the same experts ordered by their load in the current chunk, ranks 17--32 receive
at most 12 rows at $T=128$. The measured rank envelope therefore supports much smaller capacities below the first ranks.

Ranks also fix how many lanes need each capacity. Assigning an expert independently to a capacity bucket by its load
can change the number of experts in that bucket. Assigning fixed ranges of ranks to tiers keeps each tier's lane count
constant: eight ranks always occupy eight lanes, whatever their expert identities. Ranking within each SoC's expert set
keeps expert computation on its assigned device. A sort and gathers can then select the weights and token rows for
fixed-shape lanes. The small capacities depend on the measured load envelope; inputs outside it require overflow
detection and recovery (\S\ref{sec:overflow}).

The benefit also depends on executed padding. The compiler skips some empty work, so reserved capacity does not
directly determine executed work (\S\ref{sec:skip}). Smaller capacities also change buffering and memory traffic.
We therefore evaluate compiled latency as well as provisioned rows.

\subsection{Aggregation Limits the Benefit of Smaller Capacities}
\label{sec:mot-red}

After expert execution, the program must return each expert's weighted outputs to their tokens and sum the
contributions. Runtime assignment changes which expert occupies each lane, and tiers produce outputs with different
shapes. Reducing padded expert work leaves this aggregation to be performed; its storage and communication costs
determine how much of the saving becomes lower latency.

The exported combine scatters each stage's rows into a zero-filled accumulator of shape $[\mathrm{lanes},T,d]$, reduces
over each SoC's lanes and sums the four SoC partials. Each lane still reserves $T$ output rows in this accumulator even
when its expert capacity is smaller. Tiers of different widths require separate accumulators, so dense aggregation
need not become cheaper as expert padding shrinks.

Placement adds another cost. Although each expert stays on its assigned device, the compiled aggregation can
redistribute its outputs across devices; within a device, a reduction spread over cores requires exchanges between
them. Sparse aggregation must therefore also preserve locality. Section~\ref{sec:eval-own} measures these effects
with equivalent aggregation expressions that produce different compiled placements.

The interaction is visible in full-model latency. At $T=512$, replacing uniform capacity $T$ with ranked tiers lowers
Qwen3's latency by only 4\% with dense aggregation, compared with 27\% with token-owned aggregation. All four programs
use the same runtime ranking, weight access and final cross-device sum
(\S\ref{sec:eval-joint}). The capacity savings are the same, but the aggregation determines the realized speedup.

These observations motivate two requirements: assign fixed lane capacities by current load rank, and return outputs
to tokens with limited intermediate storage and communication. Section~\ref{sec:design} develops the program structure
that meets these requirements while retaining fixed shapes and placement.

\section{\sys}
\label{sec:design}

\begin{figure*}[t]
\centering
\includegraphics[width=\textwidth]{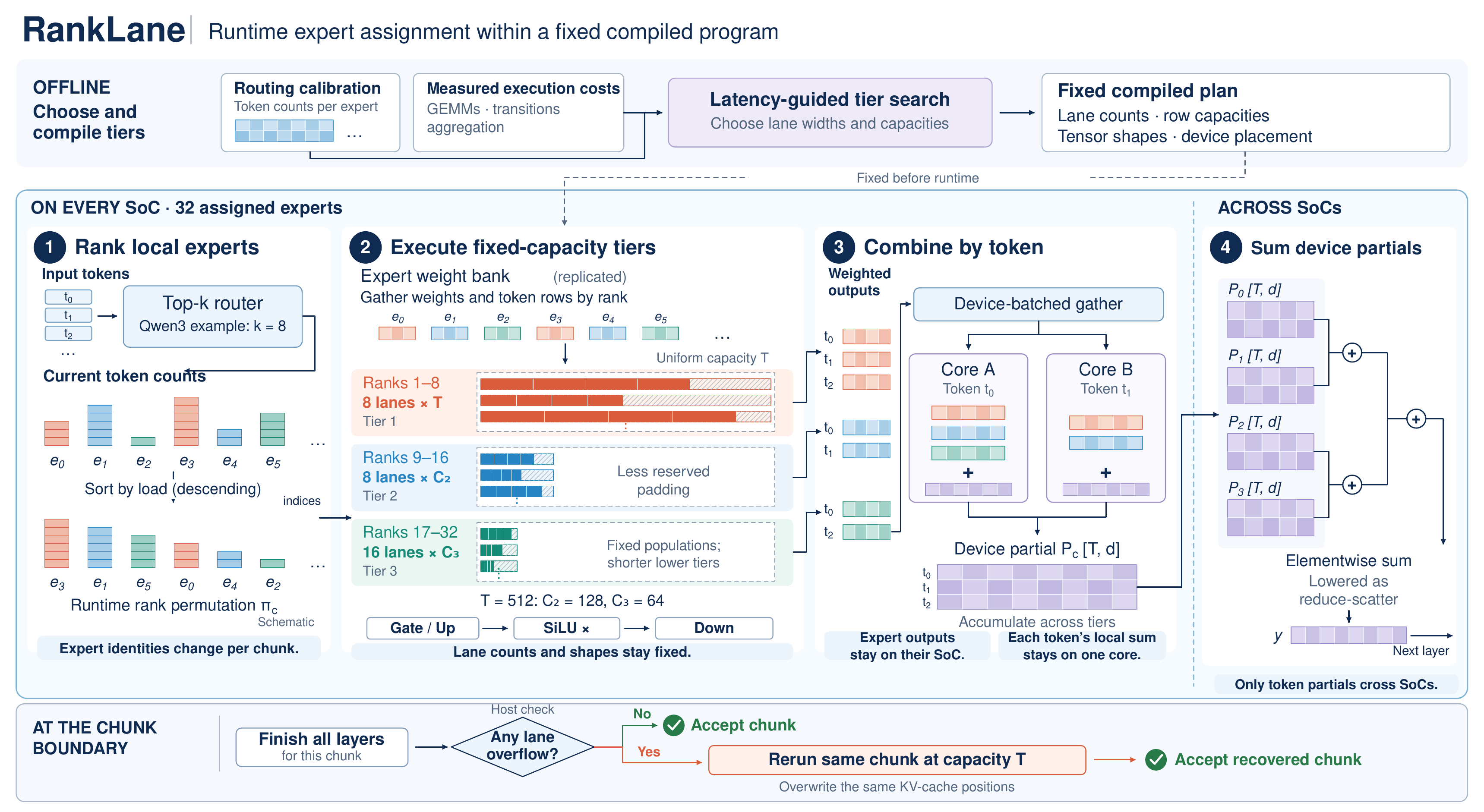}
\caption{\sys overview. Offline calibration and measured costs select fixed lane
counts and capacities. At run time, each SoC (1) ranks its experts, (2) gathers weights and tokens into fixed-capacity
lanes, and (3) combines outputs by token on local cores; (4) an elementwise sum combines device partials. After all
layers, a host check triggers a capacity-$T$ rerun if any lane overflows.}
\label{fig:overview}
\end{figure*}

\sys addresses both challenges of \S\ref{sec:mot}: it assigns experts to fixed lanes by current load rank
(\S\ref{sec:sort}), then gathers their outputs for token-local reduction (\S\ref{sec:red}). Figure~\ref{fig:overview}
shows the full pipeline within a fixed compiled graph. Calibration and measured costs select the tiers
(\S\ref{sec:calib}); overflow recovery handles loads beyond calibration (\S\ref{sec:overflow}).
Appendix~\ref{app:impl} gives the Cloud AI 100 realization.

\subsection{Adapting Expert Assignment within Fixed Tiers}
\label{sec:sort}\label{sec:tiers}\label{sec:model}

Assigning experts independently to capacity buckets would change each bucket's population, violating the fixed
lane count. \sys instead assigns fixed intervals of load ranks to tiers: busy experts receive larger lanes while
the number and shape of lanes remain constant.

\textbf{Bind capacities to ranks.} Device $c$ owns a fixed expert set $\mathcal{E}_c$, 32 experts per SoC here. The router
assigns $n_e$ tokens to expert $e$ and $N_c=\sum_{e\in\mathcal{E}_c}n_e$ to the device. Sorting these counts gives
$\pi_c$, with $n_{\pi_c(1)}\ge n_{\pi_c(2)}\ge\dots$. Tier $i$ executes one stage of $L_i$ lanes per device at capacity $C_i$,
covering ranks $a_i,\dots,a_i+L_i-1$, where $\sum_iL_i=|\mathcal{E}_c|$. Lane $j\in\{0,\dots,L_i-1\}$ serves expert
$\pi_c(a_i+j)$. Each lane executes that expert's gate, up and down GEMMs over its fixed row capacity.

\textbf{Adapt through indices.} The rank permutation selects each lane's expert weights and token table
(Figure~\ref{fig:overview}, steps 1--2). The routing computes these tables in expert order; gathers place the selected
tokens into the fixed lane inputs. Tier counts, widths, capacities, tensor shapes, weight banks and expert-to-device
placement are fixed at compilation. Only indices and data change within the compiled program. Every expert keeps one
lane and its original tokens. If $n_e$ fits that lane's capacity, the exact-arithmetic output is unchanged; Qwen3's
tested tiering variants are also bit-identical, although compilation can change summation order. A device provisions
$\sum_iL_iC_i$ rows instead of $|\mathcal{E}_c|T$.

\textbf{Why ranks.} The $r$ busiest experts each hold at least $n_{\pi_c(r)}$ assignments, and together at most $N_c$, so
\begin{equation}
n_{\pi_c(r)}\;\le\;N_c/r,\qquad N_c\;\le\;\min(k,|\mathcal{E}_c|)\,T .
\label{eq:bound}
\end{equation}
$N_c$ averages $kT|\mathcal{E}_c|/E=2T$ here and reaches at most 1.9$\times$ that in our Qwen3 256- and 512-token captures.
Lower ranks therefore admit smaller capacities, whereas fixed expert identities can peak near $T$
(\S\ref{sec:mot-pad}). Eq.~(\ref{eq:bound}) holds for any routing; our tighter empirical envelope is tested across
eight workloads (Figure~\ref{fig:rank}, \S\ref{sec:eval-gen}).

\begin{figure*}[t]
\centering
\includegraphics[width=\textwidth]{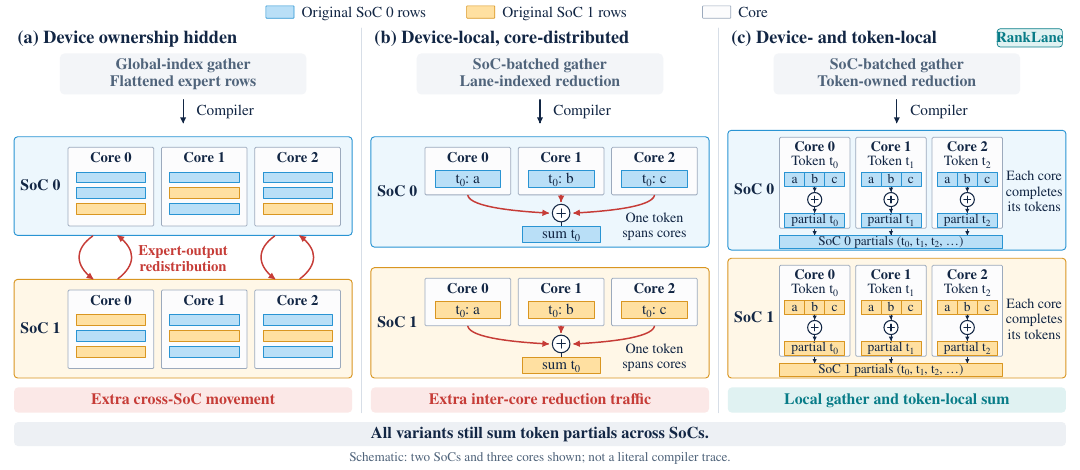}
\vskip -0.05in
\caption{Aggregation placement on the evaluated compiler. (a) Global indices cause expert-output redistribution.
(b) A SoC-batched gather preserves device locality, but lane-indexed reduction exchanges operands across cores.
(c) \sys also completes each local token sum on one core. All variants retain the final cross-SoC sum;
\S\ref{sec:eval-own} measures these placement effects.}
\label{fig:placement}
\vskip -0.1in
\end{figure*}

\subsection{Returning Outputs through Local Token Aggregation}
\label{sec:red}

The dense return path still reserves $T$ rows per lane (\S\ref{sec:mot-red}). \sys instead gathers each token's
contributions, at most $k$, and sums them locally. The graph layout must preserve locality as well as remove
the lane-dense accumulator.

\textbf{Gather contributions by token.} Let $O_c$ hold device $c$'s weighted expert outputs and $\mathrm{row}(t,j)$
address the output of token $t$'s $j$-th expert. Each device computes a token partial, followed by a sum across devices:
\begin{equation}
P_c[t]=\sum_{j:\,e_{t,j}\in\mathcal{E}_c} O_c[\mathrm{row}(t,j)],\qquad y[t]=\sum_c P_c[t],
\label{eq:agg}
\end{equation}
Figure~\ref{fig:placement} shows how the graph expression of this standard sparse combine determines device and
core locality.

\textbf{Keep expert outputs on their device.} Global indices and per-device slices cause our compiler to redistribute
expert outputs (Figure~\ref{fig:placement}(a)). \sys exposes device ownership as the gather's leading batch axis
(Figure~\ref{fig:placement}(b--c)): tier outputs reshape from $[D L_s,C_s,d]$ to $[D,L_s C_s,d]$ without moving data,
where $D$ is the device count and $L_s$ the lanes per device. Batched indices select each token's expert rows;
contributions outside the device or tier are masked.

\textbf{Sum each token's contributions on one core.} A lane-indexed reduction can stay on one device yet still exchange
operands across cores (Figure~\ref{fig:placement}(b)). \sys arranges the gathered rows as $[D,Tk,d]$ and sums over $k$ into $[D,T,d]$ partials. The compiler
splits this work by token tiles: each core gathers and sums its tokens' contributions, without a cross-core reduction
(Figure~\ref{fig:placement}(c)). Section~\ref{sec:eval-own} measures both placement effects.

\textbf{Combine device partials once.} After accumulating all tiers' local partials, the program sums
$(P_0+P_1)+(P_2+P_3)$ per token tile (Figure~\ref{fig:overview}, step 4). The compiler lowers this elementwise tree as a reduce-scatter with no
root device; the next layer's input normalization completes it. Algorithm~\ref{alg:combine} in Appendix~\ref{app:impl}
gives the complete path.

Each tier gathers $Tk$ rows per device, including masked slots, versus $L_sT$ lane-dense rows. Only $[T,d]$ partials
enter the final cross-device sum. Added tiers still pay for gathers and local partials, which tier selection must include.

\subsection{Choosing Tiers for Lower Latency}
\label{sec:calib}\label{sec:sim}

Ranking and local aggregation make smaller capacities useful, but do not determine how many tiers to compile.
More tiers follow the rank envelope more closely while adding expert stages, aggregation gathers and transitions.
We choose their widths and capacities together (Figure~\ref{fig:overview}, top): calibration bounds the observed demand at each rank, and measured
execution costs determine which groups of ranks are worth separating.

\textbf{Calibrating capacities.} A configuration $\mathcal{S}=((L_i,C_i))_{i=1}^{K}$ has $K$ tiers in rank order.
Let $m(r)$ be the largest load at rank $r$ over calibration chunks, layers and devices. Tier $i$ starts at rank $a_i$:
\begin{equation}
\begin{aligned}
a_i &= 1+\sum_{j<i}L_j, \qquad C_1=T,\\
C_i &= \min\!\left\{T,\;g\max\!\left(1,\left\lceil\frac{h\,m(a_i)}{g}\right\rceil\right)\right\},\quad i>1.
\end{aligned}
\label{eq:tier-capacity}
\end{equation}
Here $h$ is headroom and $g$ the row granularity; each lane reserves at least one quantum. Sorted counts make the first
rank the largest in its tier, so widths determine capacities. The deployed rule uses $h=1.28$ and $g=16$
(Appendix~\ref{app:calib}). The main $T=512$ search uses $g=32$ to match its profiling grid; the other searches use
$g=16$, interpolating table entries where necessary. Headroom is empirical, so overflow still requires recovery
(\S\ref{sec:overflow}).

\textbf{Account for executed padding.}\label{sec:skip}
Cloud AI 100 can skip empty work, so provisioned rows alone do not predict latency. Capacities also affect buffering,
weight loads and activation spills (Appendix~\ref{app:attrprof}). We therefore use measured compiled execution costs
to evaluate candidate tiers.

\textbf{Include the return path in the cost.} Width changes core placement and memory traffic as well as padding.
For a stage of $W$ lanes per device at capacity $C$, we measure its GEMM time $\mathrm{run}(W,C)$ and the gap to the
next stage $\mathrm{switch}(W,C)$ in complete two-layer programs with real routing and deployment settings.
Let $\sigma(j)$ identify the $j$-th tier in predicted compiler execution order. The estimated layer time is
\begin{equation}
\begin{aligned}
\widehat{\tau}(\mathcal{S})={}&\tau_{\mathrm{pre}}+\sum_{i=1}^{K}\mathrm{run}(L_i,C_i)\\
&+\sum_{j=1}^{K-1}\mathrm{switch}(L_{\sigma(j)},C_{\sigma(j)})\\
&+A(L_{\sigma(K)},C_{\sigma(K)})+\Delta(\mathcal{S},\sigma).
\end{aligned}
\label{eq:tier-cost}
\end{equation}
$\tau_{\mathrm{pre}}$ covers sort-to-first-GEMM time; $A$ covers aggregation after the last GEMM. $\Delta$ contains
fitted corrections for the first transition, later-stage slowdown and activation spills. The last stage pays no
transition. Rank order sets capacities; execution order sets costs. We predict decreasing-width execution, breaking
ties by rank order (Appendix~\ref{app:sim}).

\textbf{Searching structures.} With $N=|\mathcal{E}_c|$ experts per device, the search objective is
\begin{equation}
\begin{aligned}
\mathcal{S}^{\star}&\in\operatorname*{arg\,min}_{\mathcal{S}}\widehat{\tau}(\mathcal{S})\\
\text{subject to}\quad
&\sum_{i=1}^{K}L_i=N,\quad L_i\in\{1,\ldots,W_{\max}\},\\
&K\in\{1,\ldots,K_{\max}\},\quad C_i\text{ from Eq.~(\ref{eq:tier-capacity})}.
\end{aligned}
\label{eq:tier-search}
\end{equation}
We score 27,466 structures for $N=32$, $W_{\max}=16$ and $K_{\max}=5$, then compile leading candidates to update
spill corrections; unchecked candidates remain provisional (Algorithm~\ref{alg:tier-search}, Appendix~\ref{app:sim}).
The objective excludes recovery and does not guarantee unseen loads fit. Section~\ref{sec:eval-search} evaluates
selection quality and offline cost.

\textbf{Deployed configuration.} Qwen3 uses 8/8/16 lanes per device across chunk lengths, varying only capacities
(Table~\ref{tab:tiers}). It is among the search leaders at $T=128$ and 512; alternatives are close at $T=256$
(\S\ref{sec:eval-search}).

\begin{table}[t]
\caption{Qwen3 tiers: 8/8/16 lanes per device. Provisioned rows are $8C_1+8C_2+16C_3$, compared with $32T$ for uniform
capacity-$T$ lanes.}
\label{tab:tiers}
\centering
\small
\setlength{\tabcolsep}{5pt}
\begin{tabular}{@{}rrrr@{}}
\toprule
$T$ & $C_1/C_2/C_3$ & Tiered rows & Uniform rows \\
\midrule
128 & 128/48/16 & 1,664 & 4,096 \\
256 & 256/64/32 & 3,072 & 8,192 \\
512 & 512/128/64 & 6,144 & 16,384 \\
\bottomrule
\end{tabular}
\end{table}

\subsection{Recovering from Capacity Overflow}
\label{sec:overflow}

Future inputs can exceed the calibrated envelope. \sys reruns overflowing chunks with capacity $T$, preserving
smaller normal-path tiers while ensuring an undersized lane never produces an accepted result.

\textbf{Detect and rerun the chunk.} Detection reuses every lane's token count; the host compares it with the lane's
capacity after each chunk (Figure~\ref{fig:overview}, bottom). Recovery reruns that chunk at the same positions. Prefill writes the KV cache only at
the chunk's own positions, so the rerun overwrites every entry the failed attempt wrote, later chunks see only the rerun's
cache, and the result is that of an overflow-free run. Our recovery profile specializes capacities through the lengths
of small row inputs, sharing the normal profile's weights and cache.

\textbf{Charge recovery to latency.} For in-program recovery with overflow probability $p$, a chunk costs
$t_{\mathrm{ready}}+p\,t_{\mathrm{retry}}$ on average:
$t_{\mathrm{ready}}$
includes profile-support and host-check overhead, and $t_{\mathrm{retry}}$ includes the capacity-$T$ rerun and its
check. A separate program also pays activation and prefix replay (Appendix~\ref{app:recovery}).

\section{Evaluation}
\label{sec:eval}

We measure full-model performance and separate reduction, tiering and recovery contributions, then evaluate runtime
assignment, capacity, locality, latency-guided selection, robustness and recovery cost.

\subsection{Experimental Setup}
\label{sec:setup}

We evaluate Qwen3-30B-A3B and the text path of Gemma 4 26B-A4B on one four-SoC Cloud AI 100 Ultra card, at batch 1
with MXFP6 weights. The \emph{dense baseline} uses fixed expert assignments, static weights, capacity $T$ and dense
aggregation; it shares expert-parallel execution, head-parallel attention and tree routing scans with \sys.
\sys includes overflow detection and in-program recovery.
We measure uninstrumented full-model prefill on GSM8K chunks, using 60 timed inferences per pass after warm-up.
Pass medians are normalized across mirrored sessions through a shared reference program. Unless identified as Gemma,
ablations use Qwen3. Appendix~\ref{app:flags} gives the configurations, input selection and fitting procedure.

\subsection{End-to-End Performance}
\label{sec:eval-e2e}\label{sec:eval-contribution}

\begin{figure}[t]
\centering
\includegraphics[width=\columnwidth]{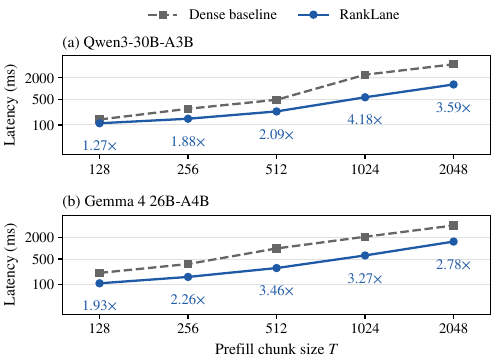}
\caption{We measure batch-one prefill with MXFP6 weights on a four-SoC Cloud AI 100 Ultra card across
128--2048-token chunks. \sys speeds up full-model prefill by 1.27--4.18$\times$ on Qwen3 and 1.93--3.46$\times$ on Gemma
over the dense baseline.}
\label{fig:eval-main}
\end{figure}

\sys prefills 1.27--4.18$\times$ faster than the dense baseline on Qwen3 and 1.93--3.46$\times$ on Gemma
(Figure~\ref{fig:eval-main}). These gains include token-owned aggregation, ranked capacities, runtime assignment
and recovery.

\begin{figure}[t]
\centering
\includegraphics[width=\columnwidth]{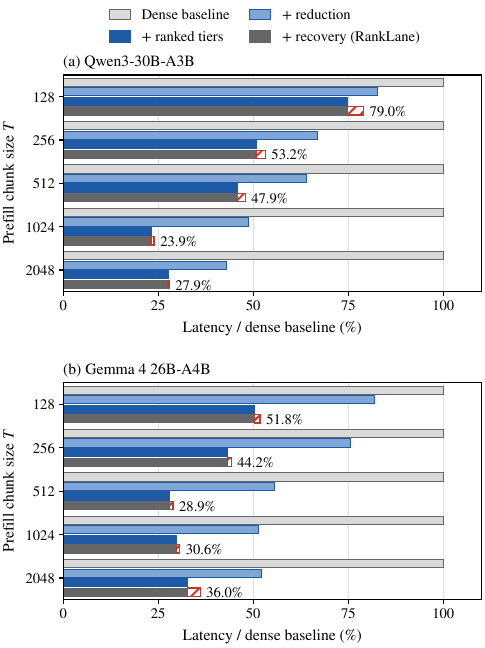}
\caption{Under the setup of Figure~\ref{fig:eval-main}, we successively add token-owned reduction, ranked tiers and
recovery support to the dense baseline. Across $T=128$--2048, reduction gives 1.21--2.32$\times$ on Qwen3 and
1.22--1.95$\times$ on Gemma over the dense baseline. Ranked tiers add a further 1.10--2.11$\times$ and
1.60--2.00$\times$, respectively, over reduction alone, before recovery overhead.}
\label{fig:cumulative}
\end{figure}

\textbf{Contribution of reduction and tiering.}
Figure~\ref{fig:cumulative} decomposes the end-to-end speedup: (1)~token-owned reduction and its
elementwise final sum, retaining static weights and capacity $T$; (2)~runtime ranking, weight gathers and deployed
tiers; and (3)~overflow detection and in-program recovery. The tiering increment includes assignment and weight
access; \S\ref{sec:eval-joint} tests the interaction.

At $T=128$--512, reduction gives 1.21--1.56$\times$ and 1.22--1.80$\times$ on Qwen3 and Gemma; ranked tiers add
1.10--1.39$\times$ and 1.63--2.00$\times$, with 4.4--5.6\% and 2.3--3.5\% recovery overhead, respectively.
At $T=512$, Gemma's 982.6$\to$546.9$\to$274.0$\to$283.7\,ms sequence saves 436\,ms through reduction and
273\,ms through tiering, then adds 10\,ms for recovery, delivering 3.46$\times$ end-to-end.
At $T=2048$, Gemma's recovery overhead grows to 10.2\%, reducing the delivered speedup to 2.78$\times$.

\subsection{Runtime Expert-to-Tier Assignment}
\label{sec:eval-abl}\label{sec:eval-pad}

We test whether calibration can fix each expert's tier for all inputs. Both alternatives in Table~\ref{tab:fixed-main}
use Qwen3's deployed 8/8/16 tier widths and capacities, the same weight gathers, token-owned reduction and final sum.
The compiled capacities stay fixed in both;
the experiment changes how experts are assigned to them.

\textbf{Assignment policies.} \emph{Calibration-fixed tiers} freeze a separate expert order for each layer and SoC.
We test orders based on mean, peak, 99th- and 99.9th-percentile calibration load and select among them using
calibration alone; the selected order is the 99th-percentile policy at all three lengths. \emph{Runtime-ranked tiers}
(\sys) instead sort the current routing counts and assign the busiest experts to the largest capacities.

\begin{table}[t]
\caption{Qwen3 assignment at matched tier capacities. In held-out trace replay, fixed assignment overflows every
chunk; runtime ranking overflows none. Attempt timings exclude recovery, required by all fixed-assignment timing runs.}
\label{tab:fixed-main}
\centering
\footnotesize
\setlength{\tabcolsep}{3pt}
\begin{tabular}{@{}rlrrr@{}}
\toprule
 $T$ & Assignment & Rows/SoC & \shortstack{Attempt\\(ms)} & \shortstack{Overflow\\chunks} \\
\midrule
128 & Calibration-fixed & 1,664 & 111.1 & 782/782 \\
 & Runtime-ranked & 1,664 & \textbf{105.4} & 0/782 \\
\midrule
256 & Calibration-fixed & 3,072 & 145.3 & 609/609 \\
 & Runtime-ranked & 3,072 & \textbf{141.7} & 0/609 \\
\midrule
512 & Calibration-fixed & 6,144 & 235.3 & 443/443 \\
 & Runtime-ranked & 6,144 & \textbf{225.5} & 0/443 \\
\bottomrule
\end{tabular}
\end{table}

\textbf{Coverage at the same budget.} We replay held-out device routing counts from 782/609/443 chunks at
$T=128/256/512$, after the exclusions in \S\ref{sec:eval-gen}. A chunk overflows if any lane in any layer exceeds
its capacity. All four calibration-fixed policies overflow every chunk at the deployed capacities; runtime ranking
overflows none of the 1,834 chunks. Both alternatives provision 59--63\% fewer rows than uniform capacity $T$.
Among the tested policies, adapting expert identities to current loads is what makes these smaller budgets cover
the held-out routing.

\textbf{Interpreting latency.} The fixed policy also overflows on each timing input, so its listed attempt must be
followed by recovery before an output can be accepted. Table~\ref{tab:fixed-main} excludes recovery support from
both programs; the complete-system results in \S\ref{sec:eval-e2e} include detection and recovery support.
Appendix~\ref{app:fixed} gives the capacity sweep, supplied-order timing control and recovery-cost estimates;
Appendix~\ref{app:tier-controls} gives additional ranking and tier-capacity controls.

\subsection{Token-Owned Aggregation}
\label{sec:eval-joint}

We test how aggregation affects the benefit of smaller expert capacities, then isolate the device and core locality
choices in \S\ref{sec:red}.

\begin{figure}[t]
\centering
\includegraphics[width=\columnwidth]{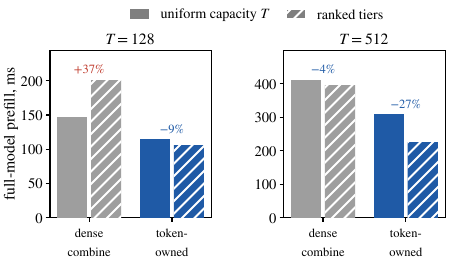}
\vskip -0.08in
\caption{Qwen3 full-model prefill under uniform and ranked capacities. Token-owned aggregation enables 9--27\%
latency reductions from tiering; dense aggregation yields a 37\% slowdown at $T=128$ and only a 4\% reduction at $T=512$.}
\label{fig:interaction}
\vskip -0.1in
\end{figure}

\textbf{Interaction with tiering.} Figure~\ref{fig:interaction} compares four Qwen3 programs at $T=128$ and 512:
uniform capacity $T$ or ranked tiers, each with dense or token-owned aggregation. All four rank experts, gather
their weights from the same bank and use the same final cross-device sum. The uniform-capacity variants here
therefore include ranking and weight gathers; Figure~\ref{fig:cumulative}'s reduction-only stage uses static weights.

Token-owned aggregation gives 1.27--1.33$\times$ speedups at uniform capacity and 1.76--1.91$\times$ with tiers.
It also makes tiering more effective: at $T=512$, tiering reduces latency by 27.1\% with token-owned aggregation,
versus 3.6\% with dense aggregation. At $T=128$, tiering increases dense-aggregation latency by 36.9\% but reduces
token-owned latency by 9.1\%. Dense aggregation retains lane-dense output buffers even when expert capacities shrink
(\S\ref{sec:mot-red}). The same capacity savings therefore yield different latency benefits depending on the return path.

\textbf{Preserving device locality.}\label{sec:eval-own}
Separate controls fix capacity $T$, static weights and the final cross-device sum to test the placements in
Figure~\ref{fig:placement}. Global row indices and explicit device slices produce the same logits as the
device-batched gather, but take 3.4--5.3$\times$ its full-model latency across $T=128/256/512$.
Instrumented layer profiles at $T=512$ show why: these expressions redistribute expert outputs, moving
1539\,MiB between SoCs per MoE layer, versus 14\,MiB when the gather exposes device ownership as a batch axis.

\textbf{Preserving core locality.} A lane-indexed reduction keeps outputs on their device but still exchanges
operands across cores. It is 11--15\% slower than token-owned aggregation across the same chunk lengths.
At $T=512$, the local sum moves 241.9\,MiB between cores, versus 1.9\,MiB when each token's local contributions
are summed on one core. Device locality alone thus leaves a substantial local communication cost.
Appendix~\ref{app:phases} gives the numerical checks and profiling protocol; these traffic measurements explain
the placement effects without decomposing full-model latency into additive phase costs.

\subsection{Latency-Guided Tier Selection}
\label{sec:eval-search}

\textbf{Tier count alone is insufficient.} In full-model Qwen3 controls, three tiers improve on two by
1.01$\times$ at $T=128$ and 1.20$\times$ at 512 (Appendix~\ref{app:tier-controls}). The benefit varies because
each tier changes execution and transition costs as well as padding. We therefore select widths and capacities
using measured execution costs rather than minimizing provisioned rows alone.

\textbf{Selecting among candidates.} Table~\ref{tab:search-qwen} in Appendix~\ref{app:sim} compares predicted and
measured Qwen3 MoE spans, fixing calibration and headroom within each chunk length. At $T=128$ and 512, the deployed
8/8/16 structure is fastest among the 8 and 21 measured configurations following the respective capacity rules.
Its $T=512$ predicted score is within 0.1\% of the search minimum. At $T=256$, the exact-fit search selects 8/16/8,
measured at 2.35\,ms versus 2.32\,ms for the fastest of five tested candidates, a 1.3\% gap.
These single-sample profiles do not resolve close alternatives.

\textbf{Full-model selection benefit.} Table~\ref{tab:search-main} tests whether the cost model improves on
Gemma's manually selected, profiled four-tier structure. At $T=512$, the search chooses 11/12/9 lanes with capacities
512/112/32 from the calibration split ($h=1.28$). In an anchored session separate from
Figure~\ref{fig:eval-main}, the selected structure
reduces full-model latency by 8.5\% despite provisioning 18.2\% more rows. It also reduces latency by 3.6\% relative to
12/11/9 at 512/96/32, calibrated on the same split with $h=1.28$.
All three programs include overflow detection and in-program recovery; none overflows on the timing chunk,
and their logits are bit-identical. The 11/12/9 widths are selected in all five calibration splits, with capacities
512/112/32 in four and 512/96/32 in one. Table~\ref{tab:search-main} fixes the primary split; the latter is a separate
sensitivity point (Appendix~\ref{app:gemma-search}).

The selected 11/12/9 structure fits one replicated weight bank and avoids short 16-lane stages that use too few
cores (Appendix~\ref{app:gemma}). We transfer these widths to the other chunk lengths and recalibrate capacities
with $h=1.28$; this does not establish a separate width optimum at each length.

The model predicts 6.65 versus 6.83\,ms per MoE layer for 11/12/9 and 12/11/9. Two-layer measurements reverse
that order (7.91 versus 7.01\,ms), while full-model timing favors the selected structure. This supports full-model
validation of cost-guided candidates, without establishing a global optimum.

\begin{table}[t]
\caption{Gemma prefill at $T=512$ with recovery support. Latency-guided tiers cut latency by 8.5\% versus manual
tiers despite 18.2\% more rows, showing that reserved capacity alone does not predict performance.}
\label{tab:search-main}
\centering
\footnotesize
\setlength{\tabcolsep}{3pt}
\begin{tabular}{@{}llrr@{}}
\toprule
Selection & Widths & Rows/SoC & ms \\
\midrule
Manual & 8/8/8/8 & 6,144 & 309.9 \\
Alternative & 12/11/9 & 7,488 & 294.2 \\
Latency-guided & 11/12/9 & 7,264 & \textbf{283.5} \\
\bottomrule
\end{tabular}
\end{table}

\begin{figure}[t]
\centering
\includegraphics[width=0.9\columnwidth]{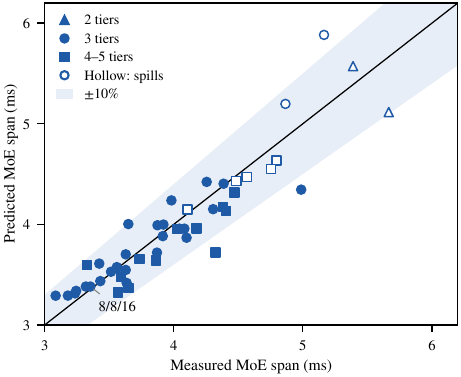}
\caption{Predicted versus measured Qwen3 MoE spans at $T=512$ for 48 programs, using the full fitted cost table.
Predictions track the measured cost differences with 4.3\% mean absolute percentage error; close candidates still
require measurement to resolve their ordering.}
\label{fig:tier-prediction}
\end{figure}

\textbf{Prediction accuracy.} Figure~\ref{fig:tier-prediction} compares predicted and measured Qwen3 MoE spans;
the full-fit mean absolute percentage error is 4.3\% across 48 programs.
Excluding each program's own cost-table observations raises it to 4.9\%; auxiliary terms still share the fitting pool.
Gemma's errors are 3.9\% on 19 independent structures and 10.9\% on 20 search-selected candidates, both excluded from
fitting. All search-candidate errors are optimistic. The models predict MoE spans, not full-model latency;
full-model validation remains necessary. Appendices~\ref{app:sim} and \ref{app:gemma-search} give validation details.

\textbf{Search cost and reuse.} Qwen3's $T=512$ cost table uses 256 grid programs, each taking about a minute to
compile and a minute to profile, with the two phases overlapped. Gemma uses 251 programs with four samples each.
Once the table is available, scoring all 27,466 Qwen3 structures takes seconds; spill checks take about a minute
per candidate. The table can score new calibration profiles for the same model, precision and compiler settings.
This concentrates device profiling on the cost grid and selected candidates, rather than every enumerated structure.

\subsection{Capacity Robustness and Recovery Cost}
\label{sec:eval-gen}

\textbf{Qwen3.} Replaying held-out device routing for 782/609/443 chunks at $T=128/256/512$ across eight workloads
gives no capacity exceedances (1,834 chunks total). The evaluation excludes calibration documents and
HumanEval/JavaScript problem overlaps; fixed policies use the same filtered set without recalibration
(Appendices~\ref{app:calib} and~\ref{app:fixed}).

\textbf{Gemma.} Executing the deployed 11/12/9 program on 430 held-out inputs at each of $T=128/256/512$ across
eight workloads produces zero overflows. Tiered and capacity-$T$ profiles give identical logits on every tested input.
Appendix~\ref{app:gemma-corpus} reports per-workload latency and output checks;
Appendix~\ref{app:gemma-device} gives calibration-split, headroom and source-holdout studies.

\textbf{Overflow recovery cost.} With an injected host overflow signal, total chunk latency is
238.3/349.9/579.0\,ms for Qwen3 and 248.1/376.2/674.3\,ms for Gemma at $T=128/256/512$:
2.1--2.5$\times$ a normal recovery-enabled chunk. These measurements include the rejected tiered attempt, both host
checks and the in-program capacity-$T$ rerun. Normal-path overhead is already included in
Figures~\ref{fig:eval-main} and~\ref{fig:cumulative}. Gemma additionally passes 180 deliberate actual-overflow
recoveries on four-chunk prompts, with accepted outputs and continuation matching direct capacity-$T$ execution
at the stressed chunk (Appendix~\ref{app:recovery}).

\section{Conclusion}
\label{sec:conclusion}

\sys enables efficient MoE prefill on accelerators with static execution plans. It binds lane capacities to current
load ranks and uses token-owned aggregation to preserve device and core locality. Calibration estimates capacity
needs, latency-guided search accounts for the execution cost of tier boundaries, and overflow recovery handles
loads beyond the calibrated envelope.

We evaluate full-model, batch-one prefill on Cloud AI 100, using MXFP6 weights and 128--2048-token
chunks. Including overflow detection and in-program recovery support, \sys achieves end-to-end prefill speedups
over a dense baseline of 1.27--4.18$\times$ on Qwen3 and 1.93--3.46$\times$ on Gemma.

\clearpage
\flushbottom
\bibliography{references}
\bibliographystyle{mlsys2025}

\raggedbottom
\section*{AI Usage Disclosure}
\label{sec:ai-usage}

We used OpenAI's ChatGPT/Codex and Anthropic's Claude Code during research and manuscript preparation.
Their use was iterative and substantial, including research planning, implementation assistance, and drafting
and revision of the manuscript.

\textbf{Ideation and literature search.}
ChatGPT/Codex helped refine the problem framing, motivation, method presentation, and formulation of tier selection.
Both ChatGPT/Codex and Claude Code were used for literature search, including identifying related systems and
accelerator architectures, summarizing relevant material, and developing comparisons with prior work.

\textbf{Experiment design and code.}
The tools helped propose and refine baselines, matched-budget comparisons, tiering and reduction ablations,
calibration and held-out evaluation protocols, and overflow-detection and recovery experiments.
ChatGPT/Codex also helped discuss candidate evaluation models and the interpretation and presentation of results.
Both tools were used to generate and revise code for experimental workflows, including program construction,
compilation, experiment orchestration, data processing, result analysis, and visualization.
Codex additionally executed build, measurement, and analysis commands through its tool interface and assisted with
debugging, reproduction instructions, and checks against experiment artifacts.

\textbf{Data preparation and results.}
AI-assisted code was used for workload preparation, tokenization, calibration/test partitioning, and analysis of
routing counts and performance logs. No study-specific dataset examples or benchmark labels were generated with
these tools. Timing and routing results were obtained from program executions and recorded experiment artifacts;
simulator predictions were computed by the cost-model implementation.

\textbf{Writing and presentation.}
ChatGPT/Codex and Claude Code were used for writing and revision. ChatGPT/Codex assistance covered the abstract,
introduction, motivation, method, evaluation, conclusion, and appendices. It provided suggested equations and
pseudocode, reviewer-style feedback, and help with figures, tables, captions, LaTeX formatting, and the separate
submission PDFs.
The authors are responsible for the final technical claims, citations, experimental results, and manuscript.

\clearpage
\raggedbottom
\appendix

\section{Experimental Setup and Compilation}
\label{app:flags}

\textbf{Hardware and models.} One Cloud AI 100 Ultra card has four SoCs with 16 cores and 32\,GB each, using SDK
1.21.6. We evaluate all 48 layers of Qwen3-30B-A3B and the full 30-layer text path of Gemma 4 26B-A4B-IT
\citep{gemmateam2026gemma4}, including embedding, final normalization and language-model head. Gemma selects
eight of 128 experts (hidden/expert widths 2816/704), adds a parallel dense MLP, and alternates five sliding-attention
layers with one global layer. Main results use batch 1, MXFP6 weights, FP16 activations and retained KV state;
ablations use Qwen3 unless identified as Gemma.

\textbf{Comparisons.} The dense baseline uses fixed expert assignments, static weights, capacity $T$ and dense output
aggregation. Both programs share expert-parallel execution, head-parallel attention and tree routing scans
(Appendix~\ref{app:impl}); Gemma also shares direct activation gathers and logit soft-capping. \sys adds token-owned
aggregation, ranked tiers and in-program recovery. The detailed tier controls already use token-owned reduction
and omit recovery support (Appendix~\ref{app:tier-controls}); they are distinct from the dense baseline.

\textbf{Timing.} Primary measurements use first-chunk GSM8K chat prefills, outside the calibration range at
$T=256,512$ for Qwen3. Mirrored passes share a uniform-capacity, token-owned anchor; a program-plus-session fit
of log latency separates program and session effects. Qwen3's $T=128/256/512$ fit uses 74--83 passes, each the
median of three 20-inference rounds after warm-up; maximum residuals are 0.63/0.88/1.7\%. Dense and reduction-only
controls use additional mirrored passes of 60 requests after two warm-ups, normalized through the same anchor.
Longer-chunk Qwen3 sessions are in Appendix~\ref{app:ladder}. Full-model latencies are uninstrumented; phase times
and traffic come from instrumented cuts. Numerical and multi-workload checks are reported separately.

Gemma uses held-out GSM8K chunks at all five lengths: four mirrored sessions, 24 passes and 60 timed requests
after two warm-ups per pass. Fit residuals are 0.58/0.57/0.67/0.57/0.42\%. Each length includes 240 injected
recoveries; single-profile, recovery-enabled and recovered logits match exactly on the timing input.

\textbf{Compilation.}
Qwen3 uses the command below with four-SoC partitioning and one specialization per chunk length (sequence $T$,
context $2T$). All compared programs use \texttt{-mdts-mos=1}. Exact per-program commands are retained in the
reproduction scripts.
{\footnotesize
\begin{minipage}{\columnwidth}
\begin{verbatim}
qaic-compile -aic-hw -aic-hw-version=ai100
  -retained-state -convert-to-fp16
  -mxfp6-matmul -aic-num-cores=16 -mos=1
  -aic-enable-depth-first
  -mdp-load-partition-config=<4 SoCs>
  -network-specialization-config=<T, 2T>
  -mdts-mos=1
\end{verbatim}
\end{minipage}}

\textbf{Gemma at $T=2048$.}\label{app:gemma-long}
Both compared programs use unwrapped cache-update positions within the retained context (exported modulus
$2^{30}$), preserving the sliding-attention mask and avoiding within-chunk overwrites beyond the 1024-token window.
The embedding table is split into four column blocks with concatenated lookups to fit the compiler's per-core
address space at context 4096; this split preserves logits bit for bit. Capacities and coverage are in
Appendix~\ref{app:gemma}.

\textbf{Scope.}\label{app:transfer}\label{sec:transfer}
\sys targets fixed expert capacities and placement with costly padding. Weight reuse and scheduling policies
\citep{awsneuronmoe} can change the preferred tiers; transfer requires new cost measurements and locality checks.
Our evidence covers two models, batch-one prefill and one compiler, without task-quality, decode, larger-batch or
other-platform evaluation. Calibration cannot guarantee unseen loads fit; reference-routing checks and final-program
executions have distinct coverage (Appendix~\ref{app:gemma}).

\section{Realization on Cloud AI 100}
\label{app:impl}

\sys transforms QEfficient ONNX exports \citep{qefficient}. Compiler cost measurements are in
Appendices~\ref{app:attrprof} and \ref{app:sim}.

\subsection{Expert Placement and Retained State}
\label{sec:ep}

Both programs keep each expert's three matrices on one SoC (\texttt{-mdts-mos=1}). Head-parallel attention uses
a leading device batch axis to colocate cache inputs and outputs as retained state. The four attention partials
are combined as $(G_0+G_1)+(G_2+G_3)$ per token tile. These choices are shared by the compared programs.

\subsection{Routing and Aggregation}
\label{sec:impl-comb}

Routing assigns compact rows through $\log_2 T$ Hillis--Steele scan steps \citep{hillis1986data}.
The dense baseline scatters the first stage's weighted outputs into a zero-filled accumulator; later stages
gather, add and scatter, and the per-SoC lane sum uses 16 token tiles.

Algorithm~\ref{alg:combine} instead gathers each token's selected expert rows with a leading device batch axis.
Other-device or other-tier entries use a safe index and zero mask because the gather does not zero-fill invalid
indices. The final elementwise sum lowers to a reduce-scatter completed by the next layer's normalization.
Appendix~\ref{app:phases} isolates device placement and core-local summation.

\begin{algorithm}[t]
\caption{Ownership-preserving token aggregation for one tier $s$ ($D$ devices)}
\label{alg:combine}
\begin{algorithmic}[1]
\REQUIRE top-$k$ experts $\mathcal{K}[t,1..k]$ of each token $t$; map $\ell(e)\mapsto(s,c,l)$ (tier, device, lane); slot table
  $\mathrm{slot}[e,t]$; tier outputs $O_s$ of shape $[D L_s,C_s,d]$, $L_s$ lanes per device
\STATE $\hat O_s\leftarrow\mathrm{reshape}(O_s,[D,L_s C_s,d])$ \COMMENT{no data movement}
\FOR{all $t\le T$, $j\le k$, devices $c$ \textbf{in parallel}}
  \STATE $e\leftarrow\mathcal{K}[t,j]$;\; $(s',c',l)\leftarrow\ell(e)$
  \STATE $m[c,t,j]\leftarrow[s'=s\wedge c'=c]$;\; $\mathit{idx}[c,t,j]\leftarrow m\,?\,l\,C_s+\mathrm{slot}[e,t]:0$
\ENDFOR
\STATE $G\leftarrow\mathrm{gather}(\hat O_s,\mathit{idx})\odot m$ \COMMENT{batched over devices: $[D,Tk,d]$}
\STATE $P_s\leftarrow\sum_j G[\cdot,\cdot,j]$ \COMMENT{per token, on its core: $[D,T,d]$}
\STATE \textbf{after all tiers:} $P\leftarrow\sum_s P_s$;\; $y\leftarrow(P_0+P_1)+(P_2+P_3)$ \COMMENT{elementwise}
\end{algorithmic}
\end{algorithm}

\subsection{Runtime Assignment and Weight Access}
\label{sec:impl-cap}

Each SoC sorts 32 counts with \texttt{TopK}; masks, scans and slot tables remain in native expert order and do not
depend on sorting. Each tier's weight gathers lower to indirect DMA from a constant containing all 128 experts.
The compiler replicates this bank across SoCs although each computes only 32 experts; Qwen3 occupies about
22.5\,GiB per SoC versus 6.7\,GiB for static weights. Replication is a lowering property, not an algorithmic
requirement. Gemma's constraints are in Appendix~\ref{app:gemma}. Each chunk length is separately specialized,
and every program returns every layer's lane counts for overflow detection.

\section{Routing Captures and Tier Calibration}
\label{app:calib}

The original capacity-calibration captures run the model's FP32 layers on the CPU and record every layer's top-8 experts. At $T=128$ we use 50
chat prompts per workload; at 512 tokens, 8 GSM8K chunks and 4 each from HumanEval, MMLU and SWE-bench Lite; the 256-token
calibration set holds 16 GSM8K chunks and both halves of each 512-token chunk (56 in total). The held-out GSM8K chunks of
the evaluation use test questions 1000--1003 (256 tokens) and 1004--1011 (512 tokens), outside the calibration range.
Table~\ref{tab:ranks} gives the largest per-SoC load at selected ranks, the input to the tier rule of \S\ref{sec:calib}.

Figure~\ref{fig:rank} instead uses the device-routing captures audited below. Each panel has 200 calibration and 200
held-out chunks from the same eight workloads. Documents are split before chunking; evaluation applies the exclusions
below. We match each workload's count across both lengths and both splits, allocating as evenly as availability permits,
then sample without replacement with seed 20261030, independently of routing loads. Calibration fixes one mean-load
expert order per layer and SoC; both assignment policies use the same held-out chunks. The capacity lines retain the original
calibration above. The diagnostic is a subset of the audited data; coverage results use the full retained set.

\textbf{Evaluation-set audit.} We reconstruct all 436 original calibration prompts from source documents and verify
their token IDs against the saved captures. Of the original 805/618/449 device-evaluation chunks, 23/5/4 contain a
document used in capacity calibration. Excluding these leaves 782/613/445; additionally grouping HumanEval problems
with their JavaScript translations leaves 782/609/443 (1,834 total). The latter set is used throughout the main-text
Qwen3 overflow comparison. It is also disjoint from the fixed policies' calibration documents under these groups.
No retained chunk exceeds the deployed capacities. The audit checks source-document identity and known translations;
it does not establish exhaustive semantic deduplication or a prospective test corpus.

\begin{table}[htbp]
\caption{Largest load of the rank-$r$ expert of a SoC over all calibration chunks, layers and SoCs.}
\label{tab:ranks}
\centering
\small
\setlength{\tabcolsep}{3.5pt}
\begin{tabular}{@{}rrrrrrrrr@{}}
\toprule
$T$ & $r=1$ & 2 & 4 & 8 & 9 & 16 & 17 & 32 \\
\midrule
128 & 127 & 122 & 105 & 33 & 32 & 14 & 12 & 3 \\
256 & 256 & 223 & 134 & 58 & 50 & 25 & 25 & 4 \\
512 & 508 & 424 & 240 & 101 & 89 & 51 & 50 & 10 \\
\bottomrule
\end{tabular}
\end{table}

\section{Tier Assignment and Compiler Effects}
\label{app:tier-controls}\label{app:ladder}\label{app:attrprof}

Table~\ref{tab:attr} fixes token-owned reduction and gathered weights. B and C use capacity $T$;
D uses 16/16 lanes at $T$ and $T/8$; E uses the deployed 8/8/16 tiers. Programs within each column share
a compilation configuration. All give bit-identical logits on the tested chunks.

\begin{table}[htbp]
\caption{Tier-assignment ablation: full-model latency (ms per chunk), with token-owned reduction in every program.
B$\to$C adds ranking; C$\to$D/E adds tiers. Dashes omit historical variants outside the common configuration.
All tiered programs here are single-profile; recovery is included in Figure~\ref{fig:cumulative}. Row budgets are per SoC.}
\label{tab:attr}
\centering
\footnotesize
\setlength{\tabcolsep}{3pt}
\begin{tabular}{@{}lrrr@{}}
\toprule
 & $T=128$ & $T=256$ & $T=512$ \\
\midrule
\multicolumn{4}{@{}l}{\textit{Qwen3-30B-A3B}} \\
B gathered, fixed order, $T$ & --- & 160.2 & 302.4 \\
C gathered, ranked, $T$ & 115.6 & 163.3 & 308.3 \\
D ranked, two tiers & 106.4 & \textbf{138.1} & 269.8 \\
E \sys, three tiers & \textbf{105.1} & --- & \textbf{224.8} \\
\midrule
C/E speedup & 1.10$\times$ & --- & 1.37$\times$ \\
Rows, B--C & 4,096 & 8,192 & 16,384 \\
Rows, D & 2,304 & 4,608 & 9,216 \\
Rows, E & 1,664 & 3,072 & 6,144 \\
\bottomrule
\end{tabular}
\end{table}

\textbf{Ranking and capacity controls.} With uniform capacity $T$ and gathered weights, B$\to$C adds ranking
and increases latency by 2.0/1.9\% at $T=256/512$. C$\to$E retains ranking and weight gathers while introducing
tiers, giving 1.10/1.37$\times$ speedups at $T=128/512$. These are whole-program latency differences.

\textbf{Executed padding.} Instrumented layer-1 profiles of two-layer $T=512$ cuts, with an einsum
attention head-group sum, show that $\mathrm{Where}(r<n_e,y,0)$ gates activation gathers, GEMMs and SiLU.
B/C execute 6.9--7.7k rows per SoC, versus 3.3--3.5k for E. Thus, executed rows differ from reserved capacity
even in the uniform-capacity programs. These trace-derived observations explain work differences but are not
an additive decomposition of full-model latency. The search also checks activation spills because buffer
placement can add DRAM traffic (Appendix~\ref{app:sim}).

\textbf{Longer chunks.} Qwen3 at $T=1024/2048$ uses one held-out GSM8K chunk and a $2T$
cache per length. Three anchored sessions (22 passes; residuals 0.21/0.94\%) establish ranked-tier timing;
three further mirrored sessions normalize dense/reduction controls through a shared
reduction anchor and recovery through single-profile \sys (residuals 0.16/0.37\%). Capacities $T,T/4,T/8$ are extrapolated,
not recalibrated; neither timing chunk overflows.

All programs share a 150\,W card limit rather than fixed clocks. Median clocks are 1018--1100\,MHz through
$T=512$; at 2048, reduction-only runs at 912\,MHz, single-profile \sys at 995\,MHz and recovery-enabled
\sys at 984\,MHz. The dense and reduction controls can throttle at the longer lengths.

\section{Aggregation Formulations and Phase Profiles}
\label{app:phases}

\label{app:locality}

\begin{figure}[htbp]
\centering
\includegraphics[width=\columnwidth]{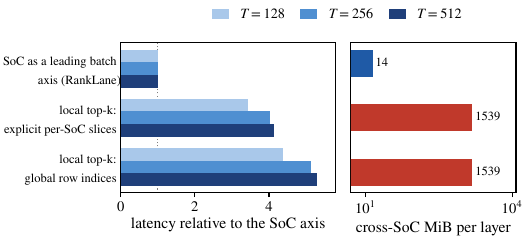}
\vskip -0.08in
\caption{Equivalent arithmetic, different placement. Three expressions of the same per-SoC token aggregation, with
bit-identical logits at uniform capacity $T$ with static weights. Left: full-model latency relative to the
layout-preserving SoC axis. Right: cross-SoC traffic inside
one MoE layer at $T=512$, from instrumented two-layer profiles.}
\label{fig:ownership}
\vskip -0.1in
\end{figure}

These controls fix capacity $T$, static weights and the final cross-device sum. The reference uses SoC-batched,
token-owned aggregation; Figure~\ref{fig:ownership} tests device placement and Table~\ref{tab:local-reduction}
tests core locality.

\textbf{Device ownership.} Figure~\ref{fig:ownership} compares global row indices, explicit SoC slices and the SoC
batch axis while holding the selected rows, summation order and final sum fixed. All three produce bit-identical
logits. Global indices and explicit slices make the compiler redistribute expert outputs: 1539\,MiB per layer
against 14\,MiB for the SoC batch axis at $T=512$. Their full-model latencies are respectively 4.4--5.3$\times$ and
3.4--4.1$\times$ higher across chunk lengths. This isolates the graph expression; redistribution can also change
scheduling, buffering and tiling.

\textbf{Token-local reduction.} The lane-indexed formulation preserves device locality but reduces each token tile
over lanes distributed across 16 cores. It is 11--15\% slower across chunk lengths. Table~\ref{tab:local-reduction}
shows why device locality alone is insufficient: cross-SoC traffic is similar, but the local sum moves far more data
between cores. Token-owned reduction gathers and sums each token's expert rows on one core. These two formulations
differ in FP16 summation order; the device-ownership variants above preserve the order exactly, while this comparison permits FP16 rounding differences.

\begin{table}[htbp]
\caption{Local-reduction ablation at $T=512$, uniform capacity $T$ and static weights. Full-model latency is
uninstrumented; traffic and local-sum busy time are from layer 1 of instrumented two-layer cuts, summed across SoCs.}
\label{tab:local-reduction}
\centering
\footnotesize
\setlength{\tabcolsep}{3pt}
\begin{tabular}{@{}lrr@{}}
\toprule
 & Lane-indexed & Token-owned \\
\midrule
Full-model latency (ms) & 401.9 & 349.3 \\
Cross-SoC traffic (MiB) & 12.1 & 14.0 \\
Local-sum inter-core traffic (MiB) & 241.9 & 1.9 \\
Local-sum busy time (core-ms) & 47.1 & 2.4 \\
\bottomrule
\end{tabular}
\end{table}

Both forms gather similar amounts of data (898/924\,MiB), but lane-indexed sums pull operands from 16 cores.
Instrumented phase gaps exceed uninstrumented per-layer gaps and cannot be added to reconstruct full-model
latency. Figure~\ref{fig:interaction}'s dense control uses one accumulator per tier width, writes the first stage
without reading zeros, and sums lanes in 16 token tiles (\S\ref{sec:eval-joint}).

\section{Tier Search: Cost Tables and Validation}
\label{app:sim}

\subsection{Cost Tables and Fitting}

\begin{figure*}[t]
\centering
\includegraphics[width=\textwidth]{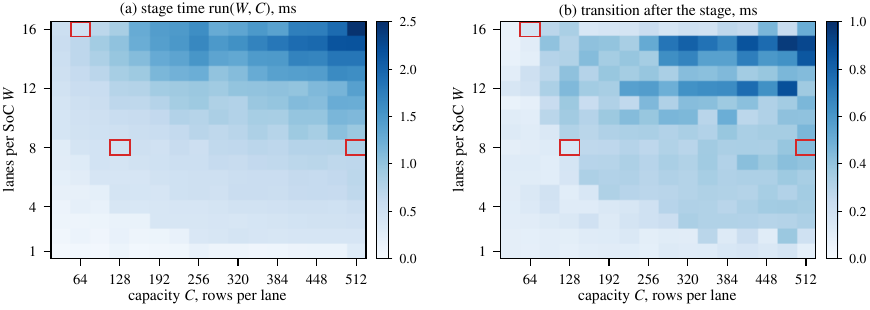}
\vskip -0.08in
\caption{Qwen3 cost table at $T=512$: (a) stage GEMM span and (b) transition gap, measured on layer 1 of
instrumented two-layer cuts; boxes mark deployed tiers. Figure~\ref{fig:tier-prediction}
compares predicted and measured MoE spans.}
\label{fig:sim}
\vskip -0.1in
\end{figure*}

Figure~\ref{fig:sim} summarizes 256 programs at $T=512$, covering $W=1,\ldots,16$ lanes and
$C=32,64,\ldots,512$ rows. Each combines one
$W\times C$ stage with 16-row filler tiers for the other experts. Layer 1 of a
two-layer cut is measured on one prompt using the third of five iterations: one instrumented sample per cell.
Stage time is the mean first-to-last-GEMM span over SoCs; transition time is the mean gap to the next stage.

Equation~(\ref{eq:tier-cost}) adds sort-to-first-GEMM work, first-transition and final-aggregation corrections,
and a spill correction. The first-transition and aggregation terms are least-squares fits over capacity sweeps
and structures with known boundary stages; the spill term uses five spilling four-/five-tier programs.
Copy counts of at least 60 trigger the additional spill correction, except when the cells already include it
(14-/16-lane first stages at capacity 512). Stages are ordered by decreasing width, ties in rank
order; this matches 44 of 48 validation programs. A compile check is needed because small capacity changes
can alter spilling, adding 0.5--0.9\,ms per layer.

\subsection{Validation Protocol}

The 48-program set includes capacity sweeps, 37 structures with two to five tiers, and a 12-layer cut.
Measurements use layer 1 of two-layer cuts, except the labeled deployed structure, which uses the median of
11 layers in the 12-layer cut. In Figure~\ref{fig:tier-prediction}, shading marks $\pm10\%$ error and hollow markers
indicate spills.
The measurement spans rank-sort start to final aggregation; starting at routing would include hoisted work
overlapping attention. Of the programs, 31 were predicted before execution and 17 supplied cost-table observations.
Figure~\ref{fig:tier-prediction} in the main text shows the full-fit predictions. Mean absolute percentage error is 4.3\%, or 4.9\%
when each program's own cell observations are excluded. Auxiliary-term
fitting still uses this pool, so the latter is not a fully independent fit. Stage context and single-sample
variability limit fine ranking: the same $16\times512$ stage takes 1.98--2.93\,ms across four programs.

\subsection{Enumeration and Calibration}

Algorithm~\ref{alg:tier-search} scores candidates, compiles a leading batch for spill checks, then rescores.
Cell costs interpolate linearly in capacity at fixed width, using the nearest endpoint outside the measured range.
Unchecked candidates omit only the additional spill term; the returned ranking retains unknown spill statuses.
Enumeration needs no device execution; the compile-check batch defaults to ten candidates and can be repeated.

\begin{algorithm}[htbp]
\caption{Tier enumeration and a batch of compile-only spill checks}
\label{alg:tier-search}
\begin{algorithmic}[1]
\REQUIRE rank maxima $m$; $T,h,g,N,W_{\max},K_{\max}$; measured tables and fitted terms; cached copy counts $Q$;
  compile-check budget $b$
\STATE $\mathcal{P}\leftarrow\varnothing$
\FOR{each ordered width tuple $(L_1,\ldots,L_K)$ satisfying the width constraints in Eq.~(\ref{eq:tier-search})}
  \STATE Derive $C_i$ by Eq.~(\ref{eq:tier-capacity}); form $\mathcal{S}=((L_i,C_i))_{i=1}^{K}$
  \STATE Set $\sigma$ by decreasing width, ties in rank order
  \STATE Score $\mathcal{S}$ by Eq.~(\ref{eq:tier-cost}), using $Q$ when available; otherwise omit the additional spill term
  \STATE Add $\mathcal{S}$, its score and spill-check status to $\mathcal{P}$
\ENDFOR
\FOR{up to $b$ lowest-scoring unchecked candidates in $\mathcal{P}$}
  \STATE Compile a two-layer candidate with deployment settings
  \STATE Cache its layer-input copy count in $Q$; classify a spill at 60 or more copies
\ENDFOR
\STATE Rescore $\mathcal{P}$ using $Q$, without double-counting spills already represented in the stage cells
\STATE \textbf{return} $\mathcal{P}$ sorted by predicted latency, retaining unknown spill statuses
\end{algorithmic}
\end{algorithm}

\begin{table}[t]
\caption{Qwen3 search inputs: maxima at ranks 1/9/17 from expanded captures (401/400/252 chunks).
Deployed capacities use the original captures in Appendix~\ref{app:calib}. Headroom rules define separate comparisons.}
\label{tab:simT}
\centering
\footnotesize
\setlength{\tabcolsep}{2pt}
\begin{tabular}{@{}rllll@{}}
\toprule
$T$ & Maxima & Deployed & $h=1.28$ & $h=1$ \\
\midrule
128 & 127/30/12 & 128/48/16 & 128/48/16 & 128/32/16 \\
256 & 254/56/26 & 256/64/32 & 256/80/48 & 256/64/32 \\
512 & 509/98/44 & 512/128/64 & 512/128/64 & 512/112/48 \\
\bottomrule
\end{tabular}

\end{table}

The $T=128,256$ tables use their respective prompts, refitting sort-to-first-GEMM and final
aggregation while retaining the $T=512$ first-gap and spill terms. Their 10/15 validation programs have mean
absolute errors of 3.2/4.1\%, including exact-fit and fixed-headroom variants.
Table~\ref{tab:search-qwen} compares only candidates under the same rule: eight at $T=128$, five at 256 and
21 at 512. Arbitrary capacity sweeps and the 12-layer result are excluded from those sets. All 27,466 structures
are scored, but selection quality is relative to the measured subset, not a global optimum.
At $T=512$, the model identifies the slower two-tier structure, but 6/6/4/16 nearly matches the deployed
structure despite its higher predicted cost. Single-sample profiles do not resolve these close alternatives.

\begin{table}[t]
\caption{Qwen3 candidate selection: predicted and measured layer-1 MoE spans (ms) in instrumented two-layer cuts.
Each group fixes calibration and $h$; $^\ast$ marks the selected candidate
($T=512$: the deployed choice, within 0.1\% of the minimum score). Capacities follow Eq.~(\ref{eq:tier-capacity});
The comparison excludes arbitrary capacity sweeps and the 12-layer cut.}
\label{tab:search-qwen}
\centering
\footnotesize
\setlength{\tabcolsep}{4pt}
\begin{tabular}{@{}rrlrr@{}}
\toprule
$T$ & $h$ & Widths & Predicted & Measured \\
\midrule
128 & 1.28 & 8/8/16$^\ast$ & 2.04 & 1.94 \\
 & & 16/16 & 2.09 & 1.98 \\
\midrule
256 & 1.00 & 8/16/8$^\ast$ & 2.38 & 2.35 \\
 & & 9/7/16 & 2.43 & 2.32 \\
 & & 8/8/16 & 2.48 & 2.39 \\
\midrule
512 & 1.28 & 8/8/16$^\ast$ & 3.38 & 3.32 \\
 & & 6/6/4/16 & 3.59 & 3.33 \\
 & & 8/8/8/8 & 3.96 & 4.03 \\
 & & 16/16 & 5.12 & 5.67 \\
\bottomrule
\end{tabular}
\end{table}

\section{Qwen3 Numerical Checks}
\label{app:acc}

Table~\ref{tab:acc} checks the programs of Figure~\ref{fig:eval-main} against FP32 CPU inference on the timing chunks.
These last-position checks do not establish task-level quality.

\begin{table}[htbp]
\caption{Relative $L_2$ error of last-position logits against FP32; \checkmark\ indicates the same next token.}
\label{tab:acc}
\centering
\footnotesize
\setlength{\tabcolsep}{4pt}
\begin{tabular}{@{}lccc@{}}
\toprule
Program & $T=128$ & $T=256$ & $T=512$ \\
\midrule
Dense baseline & 0.199 \checkmark & 0.072 \checkmark & 0.087 \checkmark \\
\sys & 0.188 \checkmark & 0.091 $\times$ & 0.096 \checkmark \\
\bottomrule
\end{tabular}
\end{table}

The tier-assignment controls produce bit-identical logits on these chunks. Changes to aggregation or
compiler placement can change FP16 summation order; at $T=256$ the leading two FP32 logits are 0.58 apart.
No lane exceeds its capacity. Recovery-enabled and recovered outputs match single-profile \sys bit for bit
(Appendix~\ref{app:recovery}).

At $T=1024,2048$, the evaluated Qwen3 programs' next token and top five agree with FP32. Across those programs,
last-position KL divergence is 0.0039--0.0048 and 0.0022--0.0031, respectively. Recovery-enabled, recovered and
single-profile \sys give identical logits. These remain single-chunk checks.

\section{Overflow Recovery Protocol and Validation}
\label{app:recovery}

\textbf{Primary cost.} Using Appendix~\ref{app:flags}'s timing protocol, we inject a host overflow signal after
an otherwise fitting tiered attempt, then check a capacity-$T$ rerun. Latency includes the rejected attempt,
both checks and rerun; it measures the recovery path, not deployment overflow frequency. Recovery-cost passes
use 60 timed requests after two warm-ups; Gemma repeats the mirrored schedule in two sessions. Normal-path
overhead includes compiler profile effects as well as host checks.

\textbf{Shared state.} Shape-specialized row indices, formed from $\mathrm{len}(\mathrm{cap\_rows})$, let normal
and capacity-$T$ profiles share weights and KV state. The host checks every lane count after each attempt.
A rerun overwrites the rejected chunk's positions; later chunks return to normal tiers. Recovery-enabled and
recovered outputs match single-profile \sys exactly on both models' timing inputs.

\begin{table}[htbp]
\caption{Additional total latency when recovery is invoked (ms), including the capacity-$T$ rerun and host checks. Adding it to the normal-path latency
in Figure~\ref{fig:eval-main} gives the recovered-chunk cost in \S\ref{sec:eval-gen}, up to rounding.}
\label{tab:recovery-primary}
\centering
\footnotesize
\setlength{\tabcolsep}{4pt}
\begin{tabular}{@{}lrrrrr@{}}
\toprule
Model / $T$ & 128 & 256 & 512 & 1024 & 2048 \\
\midrule
Qwen3 & 127.4 & 202.2 & 344.4 & 1065.9 & 2285.2 \\
Gemma & 140.1 & 214.7 & 390.6 & 1197.6 & 2476.6 \\
\bottomrule
\end{tabular}
\end{table}

\textbf{Qwen3 continuation.} Final 48-layer programs process one four-chunk GSM8K prompt per $T=128,256,512$
with a $4T$ cache, mirrored passes and ten repetitions. Injecting overflow at chunk 0, 1 or 3 preserves every
accepted output, including later chunks, bit for bit against static-capacity execution. Peak sampled additional
allocation is 22.59--22.72\,GiB per SoC relative to pre-load state. A separate resident fallback requires
activation and prefix replay, with 9.26--9.38\,s switching in these tests; the deployed path shares state instead.
A two-layer cut also provokes actual overflow in 28 (layer, lane) pairs and recovers all four chunks exactly.
Full-model Qwen3 cost measurements use injected signals.

\textbf{Gemma actual overflow.} Three math/coding/news inputs run as four $T=256$ chunks through all 30 layers.
Besides normal 256/64/16 and capacity-$T$ profiles, an undersized 256/16/4 profile forces actual overflow at chunk
0, 1 or 3. Every stressed attempt is rejected and rerun, without injected signals. Ten repetitions per input in
each mirrored pass give 180 timed recoveries. All accepted outputs and continuation match a reference that
uses capacity $T$ directly at the stressed chunk and normal tiers elsewhere. This checks retained-state recovery;
undersized profiles do not estimate deployment overflow probability, and the tests do not cover decode or broad
continuing-prefix distributions.

\begin{table}[htbp]
\caption{Gemma actual-overflow validation at $T=256$: median four-chunk prompt latency (ms), pooled over three
inputs and two mirrored passes. The direct reference uses capacity $T$ at the specified chunk, with no failed attempt.
The three-profile normal schedule takes 659.4\,ms; these stress timings are separate from the deployed two-profile results.}
\label{tab:gemma-actual-recovery}
\centering
\footnotesize
\begin{tabular}{@{}rrr@{}}
\toprule
Stressed chunk & Direct reference & With actual overflow \\
\midrule
0 & 710.5 & 863.7 \\
1 & 709.3 & 859.5 \\
3 & 706.8 & 861.7 \\
\bottomrule
\end{tabular}
\end{table}

\section{Fixed Assignments at Matched Capacity Budgets}
\label{app:fixed}

\textbf{Policies and capacity search.} We reuse the device counts of \S\ref{sec:eval-gen}: 782/608/441 calibration
chunks and 782/609/443 evaluation chunks at $T=128/256/512$, after the exclusions in Appendix~\ref{app:calib}.
Policy orders and capacity choices remain frozen from the original calibration split. Each fixed policy orders experts separately per layer
and SoC by mean, peak, 99th- or 99.9th-percentile calibration load; ties use mean load. At each capacity pair,
\emph{calibration-best} selects the policy with the fewest overflowing calibration chunks, breaking ties by excess
assignments. The search fixes tier widths at 8/8/16 and the first capacity at $T$, and enumerates
$16\le C_3\le C_2\le T$ in 16-row steps. Neither policy nor capacity selection uses evaluation overflow.

\begin{table}[htbp]
\caption{Fixed assignment versus runtime ranking. Overflow is the percentage of evaluation chunks with any lane
over capacity. Rows are per SoC. The lower panel reports minimum row budgets with zero calibration overflow and
their evaluation overflow in parentheses. The evaluation-set qualification in \S\ref{sec:eval-gen} applies.}
\label{tab:fixed-capacity}
\centering
\footnotesize
\setlength{\tabcolsep}{3pt}
\begin{tabular}{@{}lrrr@{}}
\toprule
 & $T=128$ & $T=256$ & $T=512$ \\
\midrule
\multicolumn{4}{@{}l}{Same capacities as deployed \sys} \\
Rows & 1,664 & 3,072 & 6,144 \\
Runtime rank, overflow & 0\% & 0\% & 0\% \\
Each of four fixed policies & 100\% & 100\% & 100\% \\
\midrule
\multicolumn{4}{@{}l}{Capacities selected for zero calibration overflow} \\
Calibration-best, rows & 3,840 & 7,040 & 13,056 \\
\quad (evaluation overflow) & (1.4\%) & (2.1\%) & (15.8\%) \\
Fixed 99th, rows & 4,096 & 7,680 & 15,488 \\
\quad (evaluation overflow) & (0.0\%) & (0.3\%) & (0.5\%) \\
Uniform capacity $T$, rows & 4,096 & 8,192 & 16,384 \\
\bottomrule
\end{tabular}
\end{table}

At the deployed capacities, calibration-best selects the 99th-percentile policy at all three lengths. All four fixed
policies also overflow every evaluation chunk at every other capacity pair within those row budgets. When minimizing
rows subject to zero calibration overflow, calibration-best instead selects peak order. Its smaller budgets sacrifice
evaluation coverage, especially at $T=512$ (Table~\ref{tab:fixed-capacity}). These comparisons cover the four tested
calibration policies, not all possible fixed assignments.

\textbf{Device controls.} The fixed-99th and supplied-order programs share \sys's tier capacities, weight-bank
gathers, aggregation and final sum. Both remove runtime ranking while retaining
runtime-valued gather indices. The supplied order comes from the timing chunk's FP32 routing and produces
bit-identical logits to \sys, but disagrees with the MXFP6 device order, including strictly busier experts
placed below tier boundaries. The latency difference includes placement effects and does not isolate sorting.

The supplied-order program takes 103.1/139.8/229.0\,ms at $T=128/256/512$, against the
105.4/141.7/225.5\,ms ranked control in Table~\ref{tab:fixed-main}. \sys differs by $+2.3\%$, $+1.3\%$ and
$-1.5\%$, respectively. These are whole-program differences that include tier-placement effects, rather than
an isolated measurement of sorting.

\textbf{Timing and recovery estimate.} Mirrored anchor/control passes extend the log-latency fit, with maximum
residuals 0.50/0.88/1.69\%. In this expanded fit, a separately optimized uniform-capacity recovery program
takes $L_{\mathrm{full}}=113.9/169.0/288.3$\,ms at $T=128/256/512$. It uses fixed expert assignments,
static weights and token-owned reduction; capacity $T$ guarantees no overflow.

Fixed-99th overflows on the timing inputs and is charged as a failed attempt. The estimate
$\widehat L=L_{\mathrm{attempt}}+pL_{\mathrm{full}}$ combines these single-input times with test-set overflow probability
($p=1$ for fixed-99th, $0$ for \sys). It gives 225.0/314.3/523.6\,ms for fixed-99th, or 2.13--2.32$\times$
\sys's single-profile latency. It excludes host checks, switching and prefix replay and is not a measured
recovery sequence or a corpus-wide latency average. Appendix~\ref{app:recovery}'s measured recovery paths
concern \sys only.

\section{Gemma 4: Calibration and Validation}
\label{app:gemma}

Model, compilation and timing settings are consolidated in Appendix~\ref{app:flags}.

\textbf{Tiers and capacity coverage.} The delivered program uses 11/12/9 lanes per SoC, selected at $T=512$
and transferred to other lengths. Capacities are calibrated separately with $h=1.28$ and $g=16$
(Table~\ref{tab:gemma-capacity}), using uniform-capacity device routing from 449/449/451/232/126 calibration chunks.
The corresponding held-out sets contain 430/430/430/224/120 chunks, with no offline exceedances. Splits retain
source-document groups and exclude timing-prompt documents; longer inputs concatenate groups within their workload
and split. These are retrospective partitions. Direct final-program coverage at the first three lengths is reported
below; longer-length checks cover reference routing and the timing inputs, not corpus-wide tiered execution.

\begin{table}[htbp]
\caption{Delivered Gemma capacities for 11/12/9 lanes. Rows are provisioned per SoC; uniform capacity reserves $32T$.}
\label{tab:gemma-capacity}
\centering
\footnotesize
\setlength{\tabcolsep}{4pt}
\begin{tabular}{@{}rrrr@{}}
\toprule
$T$ & $C_1/C_2/C_3$ & Tiered rows & Uniform rows \\
\midrule
128 & 128/32/16 & 1,936 & 4,096 \\
256 & 256/64/16 & 3,728 & 8,192 \\
512 & 512/112/32 & 7,264 & 16,384 \\
1024 & 1024/192/64 & 14,144 & 32,768 \\
2048 & 2048/384/128 & 28,288 & 65,536 \\
\bottomrule
\end{tabular}
\end{table}

\textbf{Width and memory constraints.} Two-layer probes show that a 16-lane stage with 16 or 32 rows uses only
four cores per SoC, serializing expert work. Qwen3's 8/8/16 layout also induces duplicate replicated MXFP6 banks,
projected to require about 36\,GiB per SoC. The selected 11/12/9 structure fits with one replicated bank.
Its benefit over a manually selected four-tier layout is measured in \S\ref{sec:eval-search}. Transferring the
selected widths does not ensure the best latency at every length: at $T=256$, the earlier four-tier complete system
took 151.8\,ms versus 161.6\,ms for the frozen transferred layout, in separate anchored fits.

\subsection{Final-Program Corpus Validation}
\label{app:gemma-corpus}

All 430 held-out inputs run at each of $T=128,256,512$ on the selected-layout program, its capacity-$T$ profile
and Dense baseline: twice per input in each of two mirrored passes, after two warm-ups, starting at position zero.
None overflows; repeated outputs match exactly. Table~\ref{tab:gemma-corpus} gives per-workload ratios of median
request latency including checks, separate from the primary single-input fit. Repetitions and lengths reuse documents.

\begin{table}[H]
\caption{Final Gemma program: per-workload dense-to-complete speedup at 11/12/9 widths. Each entry uses all held-out inputs of that workload at the given length; zero actual overflows occur.}
\label{tab:gemma-corpus}
\centering
\footnotesize
\setlength{\tabcolsep}{4pt}
\begin{tabular}{@{}lrrr@{}}
\toprule
Workload & $T=128$ & $T=256$ & $T=512$ \\
\midrule
GSM8K & 1.91$\times$ & 2.23$\times$ & 3.45$\times$ \\
HumanEval & 1.94$\times$ & 2.25$\times$ & 3.46$\times$ \\
JavaScript & 1.93$\times$ & 2.25$\times$ & 3.48$\times$ \\
MMLU & 1.91$\times$ & 2.23$\times$ & 3.45$\times$ \\
SWE-bench & 1.93$\times$ & 2.24$\times$ & 3.45$\times$ \\
Alpaca & 1.87$\times$ & 2.18$\times$ & 3.37$\times$ \\
CNN/DailyMail & 1.91$\times$ & 2.23$\times$ & 3.44$\times$ \\
Chinese XNLI & 2.01$\times$ & 2.28$\times$ & 3.44$\times$ \\
\bottomrule
\end{tabular}
\end{table}

Tiered and capacity-$T$ profiles give identical logits on every input. Dense and tiered programs agree on the next
token for 384/430, 378/430 and 369/430 inputs (89.3/87.9/85.8\%); maximum relative logit $L_2$ differences are
0.356/0.463/0.731. These differences occur without overflow. The comparison establishes neither their cause nor
task accuracy or numerical equivalence of the dense and tiered graphs.

\subsection{Device-Routing Calibration at 512 Tokens}
\label{app:gemma-device}

\textbf{Capture and splits.} The 30-layer MXFP6 static-capacity program captures 916 first-chunk prefills across
eight workloads, with $8T$ assignments per layer and no capacity truncation. Five seeds split six workloads into
random chunk halves and HumanEval/JavaScript into paired problem halves, keeping document groups disjoint.
Tests exclude calibration near-duplicates and timing/cost-table documents, leaving 451 calibration chunks and
430/429/432/427/431 test chunks. These are retrospective resplits of the earlier document pool.

\textbf{Coverage and headroom.} With 11/12/9 widths, fixed $h=1.28$ gives zero overflows in every mixed split.
Calibration-only cross-validation (CV) selects $h=1.0$ in the mixed splits and $h=1.0$--1.05 in the source holdouts.
It gives one overflow per mixed split and one across the source holdouts. All failures across
mixed splits and source holdouts concern the same HumanEval chunk, whose rank-12 load reaches 84 rows against
capacity 80. On the primary split, fixed headroom raises provisioned rows from 6,880 to 7,264 (5.6\%) and covers
this outlier. Zero calibration overflow alone is therefore insufficient for held-out coverage.

\textbf{Source holdouts.} Table~\ref{tab:gemma-source} holds out each of seven sources, pairing HumanEval with
its translation. After excluding three timing-prompt chunks, 913 chunks appear once as held-out tests.
Calibration alone freezes capacities and headroom; these holdouts reuse the mixed-split document pool and do not
constitute independent trials alongside those splits.

\begin{table}[htbp]
\caption{Gemma $T=512$, 11/12/9 widths: offline overflow chunks / tested chunks under source-held-out calibration.
CV chooses headroom using calibration only.}
\label{tab:gemma-source}
\centering
\footnotesize
\setlength{\tabcolsep}{5pt}
\begin{tabular}{@{}lrr@{}}
\toprule
Held-out source & CV headroom & Fixed $h=1.28$ \\
\midrule
GSM8K & 0/149 & 0/149 \\
HumanEval + JavaScript & 1/81 & 0/81 \\
MMLU & 0/400 & 0/400 \\
SWE-bench & 0/153 & 0/153 \\
Alpaca & 0/14 & 0/14 \\
CNN/DailyMail & 0/84 & 0/84 \\
Chinese XNLI & 0/32 & 0/32 \\
\midrule
Total & 1/913 & 0/913 \\
\bottomrule
\end{tabular}
\end{table}

\textbf{Interpretation.} Offline overflow is a lane exceeding capacity under reference-program routing, not a
recovery event. Uniform-capacity and tiered routing differ by up to six rows at a rank on the reference chunk;
that check does not bound other inputs. Final-program tests cover the primary split, not every holdout configuration.
Calibration coverage is limited to $T=512$, empty KV state and MXFP6, with small Alpaca/Chinese samples.

\textbf{Calibration size and headroom.} Five seeds draw nested subsets of 32/64/128/256/451 calibration chunks
against the same 430 tests. Every subset has one exceedance at $h=1.0$ and none at 1.28. At 1.15, three 32-chunk
subsets and one 64-chunk subset have one exceedance; larger subsets have none. At 1.28, capacities are 512/96/32
for size 32, 512/(96--112)/32 for 64--256, and 512/112/32 for 451. All 120 rows are independently reproduced from
counts and frozen capacities. Reusing tests does not add independent coverage or establish that 32 chunks suffice generally.

\subsection{Gemma Cost Model and Full-Model Tier Selection}
\label{app:gemma-search}

\textbf{Fitting and validation.} Gemma uses 251 grid programs at $T=512$: recovery-enabled two-layer cuts,
MXFP6 and four instrumented samples across two rounds, averaged per program before pooling.
All 19 independent structures and 20 search candidates are excluded from fitting. Predictions span layer-1
rank sort through aggregation, not full-model latency. Table~\ref{tab:gemma-sim-error} shows optimistic errors
for search-selected candidates; stage context and transition costs are not fully captured.

\begin{table}[htbp]
\caption{Gemma simulator errors relative to measured layer-1 MoE spans. All values are percentages;
signed error is (predicted $-$ measured) / measured.}
\label{tab:gemma-sim-error}
\centering
\footnotesize
\setlength{\tabcolsep}{4pt}
\begin{tabular}{@{}lrrr@{}}
\toprule
Validation programs & Mean abs. & Max abs. & Mean signed \\
\midrule
19 independent structures & 3.9 & 9.5 & $+1.0$ \\
20 search candidates & 10.9 & 16.8 & $-10.9$ \\
\bottomrule
\end{tabular}
\end{table}

\textbf{Full-model validation.} Table~\ref{tab:search-main}'s separate 30-layer mirrored session uses 60 prompts
per pass after two warm-ups and a static-capacity anchor (maximum residual 0.28\%). All tiered programs include
recovery, fit the timing input and give identical logits. The primary comparison uses seed 0; seed 3's 512/96/32
capacities take 280.6\,ms with 7,072 rows per SoC, a separate sensitivity point. Two-layer and full-model ordering
differs (\S\ref{sec:eval-search}), so layer-level scores require full-model validation.

\end{document}